\documentclass[a4paper]{article} 
\usepackage[
hmargin=25mm, 
vmargin=25mm,
]{geometry}
\usepackage[round]{natbib}
\usepackage{authblk}
\author[1]{Voot Tangkaratt\thanks{Contacts: voot.tangkaratt@riken.jp; bo.han@riken.jp}}
\author[1]{Bo Han}
\author[1]{Mohammad Emtiyaz Khan}
\author[1,2]{Masashi Sugiyama}
\affil[1]{RIKEN AIP, Japan}
\affil[2]{The University of Tokyo, Japan}
\renewcommand\footnotemark{}
\usepackage{hyperref}
\usepackage{url}

\usepackage[utf8]{inputenc} 
\usepackage[T1]{fontenc}    
\usepackage{booktabs}       
\usepackage{amsfonts}       
\usepackage{nicefrac}       
\usepackage{microtype}      

\makeatletter
\def\Ddots{\mathinner{\mkern1mu\raise\p@
		\vbox{\kern7\p@\hbox{.}}\mkern2mu
		\raise4\p@\hbox{.}\mkern2mu\raise7\p@\hbox{.}\mkern1mu}}
\makeatother

\usepackage{amsfonts}       
\usepackage{amsmath}
\usepackage{amssymb}
\usepackage{amsthm}
\usepackage{graphicx}
\usepackage{latexsym}
\usepackage{mathabx}
\usepackage{method}
\usepackage[noend]{algpseudocode}
\usepackage[labelformat=simple]{subcaption}

\usepackage[normalem]{ulem}
\usepackage{todonotes}
\usepackage{algorithm}
\usepackage{etoolbox}
\usepackage{wrapfig}

\theoremstyle{remark}

\def\undertilde#1{\mathord{\vtop{\ialign{##\crcr
				$\hfil\displaystyle{#1}\hfil$\crcr\noalign{\kern1.5pt\nointerlineskip}
				$\hfil\widetilde{}\hfil$\crcr\noalign{\kern1.5pt}}}}}

\newcommand{\argmin}{\mathop{\mathrm{argmin}}}
\newcommand{\argmax}{\mathop{\mathrm{argmax}}}

\newcommand{\id}{{k}}

\newcommand{\state}{\boldsymbol{\mathrm{s}}}
\newcommand{\action}{\boldsymbol{\mathrm{a}}}
\newcommand{\naction}{\boldsymbol{\mathrm{u}}}

\newcommand{\dstate}{\mathrm{d}\state}
\newcommand{\daction}{\mathrm{d}\action}
\newcommand{\dnaction}{\mathrm{d}\naction}

\newcommand{\vs}{\boldsymbol{\mathrm{s}}}
\newcommand{\va}{\boldsymbol{\mathrm{a}}}
\newcommand{\vc}{\boldsymbol{\mathrm{c}}}
\newcommand{\vu}{\boldsymbol{\mathrm{u}}}
\newcommand{\vb}{\boldsymbol{\mathrm{b}}}

\newcommand{\vz}{\boldsymbol{\mathrm{z}}}

\newcommand{\vepsilon}{\boldsymbol{\mathrm{\epsilon}}}

\newcommand{\vtheta}{\boldsymbol{\mathrm{\theta}}}
\newcommand{\vphi}{\boldsymbol{\mathrm{\phi}}}
\newcommand{\vpsi}{\boldsymbol{\mathrm{\psi}}}

\newcommand{\vomega}{\boldsymbol{\mathrm{\omega}}}
\newcommand{\vbeta}{\boldsymbol{\mathrm{\beta}}}

\newcommand{\vnu}{\boldsymbol{\mathrm{\nu}}}

\newcommand{\vC}{\boldsymbol{\mathrm{C}}}

\newcommand{\vSigma}{\boldsymbol{\mathrm{\Sigma}}}

\newcommand{\diag}{\mathrm{diag}}

\newcommand{\pig}{\pi_{\vtheta}}

\newcommand{\bI}{\boldsymbol{I}}

\newcommand{\da}{\mathrm{d}\va}

\newcommand{\sd}{{d}_{\vs}}
\newcommand{\ad}{{d}_{\va}}

\usepackage{tikz}
\usetikzlibrary{fit,positioning}
\usetikzlibrary{calc}
\usepackage{tikzscale}
\usepackage{filecontents}    

\tikzset{
	plate/.style={draw, shape=rectangle, rounded corners=0.5ex, thick,
		minimum width=3.1cm, text width=3.1cm, align=right, inner sep=10pt, inner ysep=10pt,label={[xshift=-16pt,yshift=14pt]south east:#1}}
}

\tikzset{
	platex/.style={shape=rectangle, rounded corners=0.5ex, thick,
		minimum width=3.1cm, text width=3.1cm, align=right, inner sep=10pt, inner ysep=10pt,label={[xshift=-16pt,yshift=14pt]south east:#1}}
}

\let\OrgPgfTransformScale\pgftransformscale
\renewcommand*{\pgftransformscale}[1]{%
	\gdef\ScaleFactor{#1}%
	\OrgPgfTransformScale{#1}%
}
\def\ScaleFactor{1}

\title{VILD: Variational Imitation Learning\\with Diverse-quality Demonstrations}

\date{}

\begin{document}

	\maketitle
	
	\begin{abstract}
		
		The goal of imitation learning (IL) is to learn a good policy from high-quality demonstrations. However, the quality of demonstrations in reality can be diverse, since it is easier and cheaper to collect demonstrations from a mix of experts and amateurs. IL in such situations can be challenging, especially when the level of demonstrators' expertise is unknown. We propose a new IL method called \underline{v}ariational \underline{i}mitation \underline{l}earning with \underline{d}iverse-quality demonstrations (VILD), where we explicitly model the level of demonstrators' expertise with a probabilistic graphical model and estimate it along with a reward function. We show that a naive approach to estimation is not suitable to large state and action spaces, and fix its issues by using a variational approach which can be easily implemented using existing reinforcement learning methods. Experiments on continuous-control benchmarks demonstrate that VILD outperforms state-of-the-art methods. Our work enables scalable and data-efficient IL under more realistic settings than before.
		
	\end{abstract}
	
	\section{Introduction}
	\label{section:introduction}
	
	The goal of sequential decision making is to learn a policy that makes good decisions~\citep{Puterman1994}. 
	As an important branch of sequential decision making, imitation learning (IL)~\citep{Russell1998,Schaal1999} aims to learn such a policy from demonstrations (i.e., sequences of decisions) collected from experts. However, high-quality demonstrations can be difficult to obtain in reality, since such experts may not always be available and sometimes are too costly~\citep{OsaPNBA018}. This is especially true when the quality of decisions depends on specific domain-knowledge not typically available to amateurs; e.g., in applications such as robot control~\citep{OsaPNBA018}, autonomous driving~\citep{SilverBS12}, and the game of Go~\citep{SilverEtAl2016}.
	
	In practice, demonstrations are often diverse in quality, since it is cheaper to collect them from mixed demonstrators, containing both experts and amateurs~\citep{AudiffrenVLG15}. Unfortunately, IL in such settings tends to perform poorly since low-quality demonstrations often negatively affect the performance~\citep{ShiarlisMW16,LeeCO16}. For example, demonstrations for robotics can be cheaply collected via a robot simulation~\citep{MandlekarZGBSTG18}, but demonstrations from amateurs who are not familiar with the robot may cause damages to the robot which is catastrophic in the real-world~\citep{ShiarlisMW16}. Similarly, demonstrations for autonomous driving can be collected from drivers in public roads~\citep{FridmanEtAl2017}, but these low-quality demonstrations may also cause traffic accidents.. 
	
	When the level of demonstrators' expertise is known, multi-modal IL (MM-IL) may be used to learn a good policy with diverse-quality demonstrations~\citep{LiSE17,HausmanCSSL17,WangEtAl2017}. More specifically, MM-IL aims to learn a multi-modal policy where each mode of the policy represents the decision making of each demonstrator. When knowing the level of demonstrators' expertise, good policies can be obtained by selecting modes that correspond to the decision making of high-expertise demonstrators. However, in reality it is difficult to truly determine the level of expertise beforehand. Without knowing the level of demonstrators' expertise, it is difficult to distinguish the decision making of experts and amateurs, and thus learning a good policy is quite challenging. 
	
	To overcome the issue of MM-IL, existing works have proposed to estimate the quality of each demonstration using additional information from experts~\citep{AudiffrenVLG15,Wu2019,BrownGNN19}. Specifically, \cite{AudiffrenVLG15} proposed a method that infers the quality using similarities between diverse-quality demonstrations and high-quality demonstrations, where the latter are collected in a small number from experts. In contrast,~\cite{Wu2019} proposed to estimate the quality using a small number of demonstrations with confidence scores. The value of these scores are proportion to the quality and are given by an expert. Similarly, the quality can be estimated using demonstrations that are ranked according to their relative quality by an expert~\citep{BrownGNN19}. These methods rely on additional information from experts, namely high-quality demonstrations, confidence scores, and ranking. In practice, these pieces of information can be scarce or noisy, which leads to the poor performance of these methods.
	
	In this paper, we consider a novel but realistic setting of IL where only diverse-quality demonstrations are available, while the level of demonstrators' expertise and additional information from experts are fully absent. To tackle this challenging setting, we propose a new method called variational imitation learning with diverse-quality demonstrations (VILD). The central idea of VILD is to model the level of expertise via a probabilistic graphical model, and learn it along with a reward function that represents an intention of expert's decision making. To scale up our model for large state and action spaces, we leverage the variational approach~\citep{Jordan1999}, which can be implemented using reinforcement learning (RL)~\citep{SuttonBarto1998}. 
	To further improve data-efficiency when learning the reward function, we utilize importance sampling to re-weight a sampling distribution according to the estimated level of expertise.  Experiments on continuous-control benchmarks demonstrate that VILD is robust against diverse-quality demonstrations and outperforms existing methods significantly. Empirical results also show that VILD is a scalable and data-efficient method for realistic settings of IL.
	
	\section{Related Work}
	\label{section:related_work}
	In this section, we firstly discuss a related area of supervised learning with diverse-quality data. Then, we discuss existing IL methods that use the variational approach.
	
	\vspace{-2mm}
	\paragraph{Supervised learning with diverse-quality data.}		
	In supervised learning, diverse-quality data has been studied extensively under the setting of classification with noisy label~\citep{Angluin1988}. This classification setting assumes that human labelers may assign incorrect class labels to training inputs. With such labelers, the obtained dataset consists of high-quality data with correct labels and low-quality data with incorrect labels. To handle this challenging setting, many methods were proposed~\citep{RaykarYZVFBM10,Nagarajan2013,HanYYNXHTS18}.
	The most related methods to ours are probabilistic modeling methods, which aim to infer correct labels and the level of labeler's expertise~\citep{RaykarYZVFBM10,KhetanLA18}.
	Specifically, \cite{RaykarYZVFBM10} proposed a method based on a two-coin model which enables estimating the correct labels and level of expertise. Recently,~\cite{KhetanLA18} proposed a method based on weighted loss functions, where the weight is determined by the estimated labels and level of expertise. 
	
	Methods for supervised learning with diverse-quality data may be used to learn a policy in our setting. However, they tend to perform poorly due to the issue of compounding error~\citep{Ross10a}. Specifically, supervised learning methods generally assume that data distributions during training and testing are identical. However, data distributions during training and testing are different in IL, since data distributions depend on policies~\citep{NgR00}. A discrepancy of data distributions causes compounding errors during testing, where prediction errors increase further in future predictions. Due to the issue of compounding error, supervised-learning-based methods often perform poorly in IL~\citep{Ross10a}. The issue becomes even worse with diverse-quality demonstrations, since data distributions of different demonstrators tend to be highly different. For these reasons, methods for supervised learning with diverse-quality data is not suitable for IL.  
	
	\vspace{-2mm}
	\paragraph{Variational approach in IL.}
	The variational approach~\citep{Jordan1999} has been previously utilized in IL to perform MM-IL and reduce over-fitting. Specifically, MM-IL aims to learn a multi-modal policy from diverse demonstrations collected by many experts~\citep{LiSE17}, where each mode of the policy represents decision making of each expert\footnote{We emphasize that diverse demonstrations are different from diverse-quality demonstrations. Diverse demonstrations are collected by experts who execute equally good policies, while diverse-quality demonstrations are collected by mixed demonstrators; The former consists of demonstrations that are equally high-quality but diverse in behavior, while the latter consists of demonstrations that are diverse in both quality and behavior.}. A multi-modal policy is commonly represented by a context-dependent policy, where each context represents each mode of the policy. The variational approach has been used to learn a distribution of such contexts, i.e., by learning a variational auto-encoder~\citep{WangEtAl2017} and by maximizing a variational lower-bound of mutual information~\citep{LiSE17,HausmanCSSL17}. Meanwhile, variational information bottleneck (VIB)~\citep{alemi2017} has been used to reduce over-fitting in IL~\citep{peng2018variational}. Specifically, VIB aims to compress information flow by minimizing a variational bound of mutual information. This compression filters irrelevant signals, which leads to less over-fitting. Unlike these existing works, we utilize the variational approach to aid computing integrals in large state-action spaces, and do not use a variational auto-encoder or a variational bound of mutual information. 
	
	\section{IL from Diverse-quality Demonstrations and its Challenge}
	\label{section:background}
	Before delving into our main contribution, we first give the minimum background about RL and IL. Then, we formulate a new setting of IL with diverse-quality demonstrations, discuss its challenge, and reveal the deficiency of existing methods.
	
	\vspace{-2mm}
	\paragraph{Reinforcement learning.}
	Reinforcement learning (RL)~\citep{SuttonBarto1998} aims to learn an optimal policy of a sequential decision making problem, which is often mathematically formulated as a Markov decision process (MDP)~\citep{Puterman1994}. We consider a finite-horizon MDP with continuous state and action spaces defined by a tuple $\mathcal{M} = (\mathcal{S}, \mathcal{A}, p(\vs'|\vs,\va), p_1(\vs_1), r(\vs,\va) )$ with a state $\state_t \in \mathcal{S} \subseteq \mathbb{R}^{\sd}$, an action $\action_t \in \mathcal{A} \subseteq \mathbb{R}^{\ad}$, an initial state density $p_1(\state_1)$, a transition probability density $p(\state_{t+1}|\state_t,\action_t)$, and a reward function $r: \mathcal{S} \times \mathcal{A} \mapsto \mathbb{R}$, where the subscript $t \in \{1, \dots, T \}$ denotes the time step. 
	A sequence of states and actions, $(\vs_{1:T}, \va_{1:T})$, is called a trajectory.
	A decision making of an agent is determined by a policy function $\pi(\va_t|\vs_t)$, which is a conditional probability density of action given state. RL seeks for an optimal policy $\pi^\star(\va_t|\vs_t)$ which maximizes the expected cumulative reward, i.e., $
	\pi^\star = \argmax_{\pi}\mathbb{E}_{p_{\pi}(\vs_{1:T}, \va_{1:T})} [ \Sigma_{t=1}^T r(\vs_t,\va_t) ]$, 
	where $p_{\pi}(\vs_{1:T}, \va_{1:T}) = p_1(\vs_1) \Pi_{t=1}^T p(\state_{t+1}|\vs_t,\va_t) \pi(\va_{t}|\vs_t)$ is a trajectory probability density induced by $\pi$. RL has shown great successes recently, especially when combined with deep neural networks~\citep{MnihEtAl2015,SilverEtAl2017}. However, a major limitation of RL is that it relies on the reward function which may be unavailable in practice~\citep{Russell1998}.
	
\begin{figure}[t]
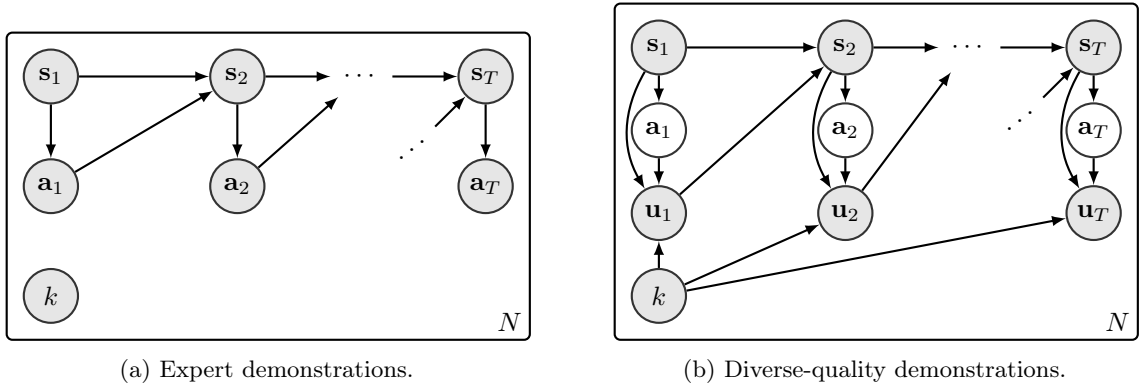

		\centering	
		\hfill
		\begin{subfigure}[b]{0.48\linewidth}
			\centering
			\includegraphics[width=0.9\linewidth]{figures/pgm_irl2.tikz}
			\subcaption{Expert demonstrations.}
			\label{figure:pgm_irl2}
		\end{subfigure}	
		\hfill
		\begin{subfigure}[b]{0.48\linewidth}
			\centering
			\includegraphics[width=0.9\linewidth]{figures/pgm_cirl.tikz}
			\subcaption{Diverse-quality demonstrations.}
			\label{figure:pgm_cirl}
		\end{subfigure}	
		\hfill
		\caption{Graphical models describe expert demonstrations and diverse-quality demonstrations.
			Shaded and unshaded nodes indicate observed and unobserved random variables, respectively. Plate notations indicate that the sampling process is repeated for $N$ times. $\vs_t \in \mathcal{S}$ is a state with transition densities $p(\vs_{t+1}|\vs_t,\va_t)$, $\va_t \in \mathcal{A}$ is an action with density $\pi^\star(\va_t|\vs_t)$, $\vu_t \in \mathcal{A}$ is a noisy action with density $p(\vu_t|\vs_t,\va_t,k)$, and $k \in \{1,\dots,K\}$ is an identification number with distribution $p(k)$.}
		\label{figure:pgm}
	\end{figure}

	\vspace{-2mm}
	\paragraph{Imitation learning.}
	To address the above limitation of RL, imitation learning (IL) was proposed~\citep{Schaal1999,NgR00}. Without using the reward function, IL aims to learn the optimal policy from demonstrations that encode information about the optimal policy. A common assumption in most IL methods is that, demonstrations are collected by $K \geq 1$ demonstrators who execute actions $\va_t$ drawn from $\pi^\star(\va_t|\vs_t)$ for every states $\vs_t$. A graphic model describing this data collection process is depicted in Figure~\ref{figure:pgm_irl2}, where a random variable $k \in \{1,\dots,K\}$ denotes each demonstrator's identification number and $p(k)$ denotes the probability of collecting a demonstration from the $k$-th demonstrator. Under this assumption, demonstrations $\{ (\state_{1:T}, \action_{1:T}, k)_n  \}_{n=1}^N$ (i.e., observed random variables in Figure~\ref{figure:pgm_irl2}) are called \emph{expert demonstrations} and are regarded to be drawn independently from a probability density $p^\star(\vs_{1:T}, \va_{1:T}) p(k) = p(k) p_1(\vs_1) \Pi_{t=1}^T p(\state_{t+1}|\vs_t,\va_t) \pi^\star(\va_{t}|\vs_t) $. We note that the variable $k$ does not affect the trajectory density $p^\star(\vs_{1:T}, \va_{1:T})$ and can be omitted. In this paper, we assume a common assumption that $p_1(\vs_1)$ and $p(\state_{t+1}|\vs_t,\va_t)$ are unknown but we can sample states from them. 
	
	IL has shown great successes in benchmark settings~\citep{HoE16,FuEtAl2018,peng2018variational}. However, practical applications of IL in the real-world is relatively few~\citep{schroecker2018generative}. One of the main reasons is that most IL methods aim to learn with expert demonstrations. In practice, such demonstrations are often too costly to obtain due to a limited number of experts, and even when we obtain them, the number of demonstrations is often too few to accurately learn the optimal policy~\citep{AudiffrenVLG15,Wu2019,BrownGNN19}.

	\vspace{-2mm}
	\paragraph{New setting: Diverse-quality demonstrations.}
	To improve practicality, we consider a new problem called \emph{IL with diverse-quality demonstrations}, where demonstrations are collected from demonstrators with different level of expertise. Compared to expert demosntrations, diverse-quality demonstrations can be collected more cheaply, e.g., via crowdsourcing~\citep{MandlekarZGBSTG18}. The graphical model in Figure~\ref{figure:pgm_cirl} depicts the process of collecting such demonstrations from $K > 1$ demonstrators. Formally, we select the $k$-th demonstrator for demonstrations according to a probability distribution $p(k)$. After selecting $k$, for each time step $t$, the $k$-th demonstrator observes state $\vs_t$ and samples action $\va_t$ using the optimal policy $\pi^\star(\va_t|\vs_t)$. However, the demonstrator may not execute $\va_t$ in the MDP if this demonstrator is not expertised. Instead, he/she may sample an action $\vu_t \in \mathcal{A}$ with another probability density $p(\vu_t|\vs_t, \va_t, k)$ and execute it. Then, the next state $\vs_{t+1}$ is observed with a probability density $p(\vs_{t+1}|\vs_t,\vu_t)$, and the demonstrator continues making decision until time step $T$. We repeat this process for $N$ times to collect \emph{diverse-quality demonstrations} $\mathcal{D}_{\mathrm{d}} = \{ (\state_{1:T}, \naction_{1:T}, k)_n \}_{n=1}^N$. These demonstrations are regarded to be drawn independently from a probability density 
	\begin{align}
	p_{\mathrm{d}}( \state_{1:T}, \naction_{1:T} | k) p(k) 
	= p(k) p(\state_1) \prod_{t=1}^T p_1(\state_{t+1}|\state_t, \naction_t) \int_{\mathcal{A}} \pi^\star(\va_t|\vs_t) p(\naction_t|\state_t,\action_t, k) \daction_t.
	\end{align}
	We refer to $p(\vu_t|\vs_t, \va_t, k)$ as a noisy policy of the $k$-th demonstrator, since it is used to execute a noisy action $\vu_t$. Our goal is to learn the optimal policy $\pi^\star$ using diverse-quality demonstrations $\mathcal{D}_{\mathrm{d}}$. 
	

	\vspace{-2mm}
	\paragraph{The deficiency of existing methods.}
	We conjecture that existing IL methods are not suitable to learn with diverse-quality demonstrations according to $p_{\mathrm{d}}$. Specifically, these methods always treat observed demonstrations as if they were drawn from $p^\star$. By comparing $p^\star$ and $p_{\mathrm{d}}$, we can see that existing methods would learn $\pi(\vu_t|\vs_t)$ such that $\pi(\vu_t|\vs_t) \approx \Sigma_{k=1}^K p(k) \int_{\mathcal{A}} \pi^\star(\va_t|\vs_t) p(\vu_t|\vs_t,\va_t,k) \da_t $. In other words, they learn a policy that averages over decisions of all demonstrators. This would be problematic when amateurs are present, as averaged decisions of all demonstrators would be highly different from those of all experts. Worse yet, state distributions of amateurs and experts tend to be highly different, which often leads to unstable learning. For these reasons, we believe that existing methods tend to learn a policy that achieves average performances and are not suitable for handling the setting of diverse-quality demonstrations. 
	
	\section{VILD: A Robust Method for Diverse-quality Demonstrations}
	
	This section describes VILD, namely a robust method for tackling the challenge from diverse-quality demonstrations. Specifically, we build a probabilistic model that explicitly describes the level of demonstrators' expertise and a reward function (Section~\ref{section:model}), and estimate its parameters by a variational approach (Section~\ref{section:variational}), which can be implemented by using RL (Section~\ref{section:model_assumption}). We also improve data-efficiency by using importance sampling (Section~\ref{section:is}). Mathematical derivations are provided in Appendix~\ref{appendix:proof}.
	
	\vspace{-1mm}
	\subsection{Model describing diverse-quality demonstrations}
	\label{section:model}
	\vspace{-1mm}
	
	This section presents a model which enables estimating the level of demonstrators' expertise. We first describe a naive model, whose parameters can be estimated trivially via supervised learning, but suffers from the issue of compounding error.
	Then, we describe our proposed model, which avoids the issue of the naive model by learning a reward function.
	
	\vspace{-2mm}
	\paragraph{Naive model.} 
	Based on $p_{\mathrm{d}}$, one of the simplest models to handle diverse-quality demonstrations is $
	p_{\vtheta, \vomega}( \state_{1:T}, \naction_{1:T}, k) = p(k) p(\state_1) \Pi_{t=1}^T p(\state_{t+1}|\state_t, \naction_t) \int_{\mathcal{A}} \pi_{\vtheta}(\va_t|\vs_t) p_{\vomega}(\naction_t|\state_t,\action_t, k) \daction_t$, 
	where $\vtheta$ and $\vomega$ are real-valued parameter vectors. These parameters can be learned by e.g., minimizing the Kullback-Leibler (KL) divergence from the data distribution to the model: $\min_{\vtheta, \vomega}  \mathrm{KL}(p_\mathrm{d}(\vs_{1:T},\vu_{1:T}|k)p(k) || p_{\vtheta, \vomega}(\vs_{1:T},\vu_{1:T},k) )$. This naive model can be regarded as a regression-extension of the two-coin model proposed by~\cite{RaykarYZVFBM10} for classification with noisy label. As discussed previously in Section~\ref{section:related_work}, such a model suffers from the issue of compounding error and is not suitable for our IL setting.
	
	\vspace{-1mm}
	\paragraph{Proposed model.} 
	
	To avoid the issue of compounding error, our method utilizes the inverse RL~(IRL) approach~\citep{NgR00}, where we aim to learn a reward function from diverse-quality demonstrations\footnote{We emphasize that IRL~\citep{NgR00} is different from RL, since RL learns an optimal policy from a known reward function.}. IL problems can be solved by a combination of IRL and RL, where we learn a reward function by IRL and then learn a policy from the reward function by RL. This combination avoids the issue of compounding error, since the policy is learned by RL which generalizes to states not presented in demonstrations. 
	
	Specifically, our proposed model is based on a model of maximum entropy IRL (MaxEnt-IRL)~\citep{ZiebartEtAl2010}.
	Briefly speaking, MaxEnt-IRL learns a reward function from expert demonstrations by using a model $p_{\vphi}(\vs_{1:T}, \va_{1:T}) ~\propto~ p(\state_1) \Pi_{t=1}^T p_1(\state_{t+1}|\state_t, \action_t)  \exp ( r_{\vphi}(\state_t,\va_t) )$. Based on this model, we propose to learn the reward function \emph{and} the level of expertise by a model
	\begin{align}
	p_{\vphi,\vomega}(\state_{1:T}, \naction_{1:T}, k) 
	= p(k) p_1(\state_1) \prod_{t=1}^T p(\state_{t+1}|\state_t, \naction_t) \int_{\mathcal{A}} \exp \left( {r_{\vphi}(\state_t,\action_t)}{}  \right) p_{\vomega}(\naction_t|\state_t,\action_t, k) \daction_t / Z_{\vphi,\vomega}, 
	\end{align}
	where $\vphi$ and $\vomega$ are parameters of the model and $Z_{\vphi,\vomega}$ is the normalization term. 
	By comparing the proposed model $p_{\vphi,\vomega}(\state_{1:T}, \naction_{1:T}, k)$ to the data distribution $p_{\mathrm{d}}$, the reward parameter $\vphi$ should be learned so that the cumulative rewards is proportion to a probability density of actions given by the optimal policy, i.e., $\exp(\Sigma_{t=1}^T r_{\vphi}(\vs_t,\va_t) ) ~\propto~ \Pi_{t=1}^T \pi^\star(\va_t|\vs_t)$. In other words, the cumulative rewards are large for trajectories induced by the optimal policy $\pi^\star$. Therefore, $\pi^\star$ can be learned by maximizing the cumulative rewards.
	Meanwhile, the density $p_{\vomega}(\naction_t|\state_t,\action_t, k)$ is learned to estimate the noisy policy $p(\naction_t|\state_t,\action_t, k)$. 
	In the remainder, we refer to $\vomega$ as an expertise parameter.

	To learn the parameters of this model, we propose to minimize the KL divergence from the data distribution to the model: $\min_{\vphi,\vomega}\mathrm{KL}(p_{\mathrm{d}}(\state_{1:T}, \naction_{1:T}| k)p(k)||p_{\vphi,\vomega}(\state_{1:T}, \naction_{1:T}, k))$. By rearranging terms and ignoring constant terms, minimizing this KL divergence is equivalent to solving an optimization problem $\max_{\vphi, \vomega} f(\vphi, \vomega) - g(\vphi, \vomega)$, where $f(\vphi, \vomega) = \mathbb{E}_{p_{\mathrm{d}}(\state_{1:T}, \naction_{1:T}| k)p(k)} [ \Sigma_{t=1}^T \log ( \int_{\mathcal{A}} \exp( {r_{\vphi}(\state_t, \action_t)}{} ) p_{\vomega}(\naction_t|\state_t,\action_t,k) \daction_t ) ]$ and $g(\vphi, \vomega) = \log Z_{\vphi,\vomega}$. To solve this optimization, we need to compute the integrals over both state space $\mathcal{S}$ and action space $\mathcal{A}$. Computing these integrals is feasible for small state and action spaces, but is infeasible for large state and action spaces. 
	To scale up our model to MDPs with large state and action spaces, we leverage a variational approach in the followings.
	
	\vspace{-1mm}
	\subsection{Variational approach for parameter estimation}
	\label{section:variational}
	\vspace{-1mm}
	
	The central idea of the variational approach is to lower-bound an integral by the Jensen inequality and a variational distribution~\citep{Jordan1999}. The main benefit of the variational approach is that the integral can be indirectly computed via the lower-bound, given an optimal variational distribution.
	However, finding the optimal distribution often requires solving a sub-optimization problem. 
	
	Before we proceed, notice that $f(\vphi, \vomega) - g(\vphi, \vomega)$ is not a joint concave function of the integrals, and this prohibits using the Jensen inequality. However, we can use the Jensen inequality to separately lower-bound $f(\vphi, \vomega)$ and $g(\vphi, \vomega)$, since they are concave functions of their corresponding integrals. Specifically, let $l_{\vphi,\vomega}(\vs_t,\va_t,\vu_t,k) = r_{\vphi}(\vs_t,\va_t) + \log p_{\vomega}(\vu_t|\vs_t,\va_t,k)$. By using a variational distribution $q_{\vpsi}(\va_t|\vs_t,\vu_t, k)$ with parameter $\vpsi$, we obtain an inequality $f(\vphi, \vomega) \geq \mathcal{F}(\vphi, \vomega, \vpsi)$, where 
	\begin{align}
	\mathcal{F}(\vphi, \vomega, \vpsi)
	&= \mathbb{E}_{p_{\mathrm{d}}(\state_{1:T}, \naction_{1:T}| k)p(k)} \left[ \sum_{t=1}^T \mathbb{E}_{q_{\vpsi}(\va_t|\vs_t,\vu_t, k)} \left[ l_{\vphi,\vomega}(\vs_t,\va_t,\vu_t,k) \right] + H_t(q_{\vpsi}) \right], 
	\end{align}
	and $H_t(q_{\vpsi}) = - \mathbb{E}_{q_{\vpsi}(\va_t|\vs_t,\vu_t, k)} \left[ \log q_{\vpsi}(\va_t|\vs_t,\vu_t, k) \right]$. It is trivial to verify that the equality $f(\vphi, \vomega) = \max_{\vpsi}\mathcal{F}(\vphi, \vomega, \vpsi)$ holds ~\citep{murphy2013}, where the maximizer $\vpsi^\star$ of the lower-bound yields $q_{\vpsi^\star}(\va_t|\vs_t,\vu_t, k) ~\propto~ \exp(l_{\vphi,\vomega}(\vs_t,\va_t,\vu_t,k))$. 
	Therefore, the function $f(\vphi, \vomega)$ can be substituted by $\max_{\vpsi} \mathcal{F}(\vphi, \vomega, \vpsi)$.
	Meanwhile, by using a variational distribution $q_{\vtheta}(\action_t, \vu_t| \state_t, k)$ with parameter $\vtheta$, we obtain an inequality  $g(\vphi, \vomega) \geq \mathcal{G}(\vphi, \vomega, \vtheta)$, where 
	\begin{align}
	\mathcal{G}(\vphi,\vomega,\vtheta) 
	&= \mathbb{E}_{\widetilde{q}_{\vtheta}(\state_{1:T}, \naction_{1:T}, \action_{1:T}, k) }	\left[ 
	\sum_{t=1}^T l_{\vphi,\vomega}(\vs_t,\va_t,\vu_t,k) - \log q_{\vtheta}(\action_t, \vu_t | \state_t, k ) \right], \label{eq:G}
	\end{align} 
	and $\widetilde{q}_{\vtheta}(\vs_{1:T},\vu_{1:T},\va_{1:T}, k) = p(k) p_1(\state_1) \Pi_{t=1}^T p(\state_{t+1}|\state_t, \naction_t) q_{\vtheta}(\action_t, \vu_t| \state_t, k)$.
	The lower-bound $\mathcal{G}$ resembles the maximum entropy RL (MaxEnt-RL)~\citep{ZiebartEtAl2010}. By using the optimality results of MaxEnt-RL~\citep{Levine2018}, we have an equality $g(\vphi,\vomega) = \max_{\vtheta} \mathcal{G}(\vphi,\vomega,\vtheta)$. 
	Therefore, the function $g(\vphi,\vomega)$ can be substituted by $\max_{\vtheta} \mathcal{G}(\vphi,\vomega,\vtheta)$. 
	
	By using these lower-bounds, we have that $\max_{\vphi, \vomega} f(\vphi, \vomega) - g(\vphi, \vomega) = \max_{\vphi, \vomega, \vpsi}  \mathcal{F}(\vphi, \vomega, \vpsi) - \max_{\vtheta}\mathcal{G}(\vphi, \vomega, \vtheta) = \max_{\vphi, \vomega, \vpsi} \min_{\vtheta} \mathcal{F}(\vphi, \vomega, \vpsi) - \mathcal{G}(\vphi, \vomega, \vtheta)$. Solving the max-min problem is often feasible even for large state and action spaces, since $\mathcal{F}(\vphi, \vomega, \vpsi)$ and $\mathcal{G}(\vphi, \vomega, \vtheta)$ are defined as an expectation and can be optimized straightforwardly. Nevertheless, in practice, we represent the variational distributions by parameterized functions, and solve the sub-optimization (w.r.t.~$\vpsi$ and~$\vtheta$) by stochastic optimization methods. 
	However, in this scenario, the equalities $f(\vphi, \vomega) = \max_{\vpsi}\mathcal{F}(\vphi, \vomega, \vpsi)$ and $g(\vphi,\vomega) = \max_{\vtheta} \mathcal{G}(\vphi,\vomega,\vtheta)$ may not hold for two reasons. First, the optimal variational distributions may not be in the space of our parameterized functions. Second, stochastic optimization methods may yield local solutions. 
	Nonetheless, when the variational distributions are represented by deep neural networks, the obtained variational distributions are often reasonably accurate and the equalities approximately hold~\citep{RanganathGB14}.
	
	
	\vspace{-1mm}
	\subsection{Model specification}
	\label{section:model_assumption}
	\vspace{-1mm}
	
	In practice, we are required to specify models for $q_{\vtheta}(\va_t,\vu_t|\vs_t,k)$ and $p_{\vomega}(\naction_t|\state_t,\action_t, k)$. We propose to use $q_{\vtheta}(\va_t,\vu_t|\vs_t,k) = q_{\vtheta}(\va_t|\vs_t) \mathcal{N}(\vu_t | \va_t, \vSigma)$ and $p_{\vomega}(\naction_t|\state_t,\action_t, k) = \mathcal{N}(\vu_t | \va_t, \vC_{\vomega}(k))$. As shown below, the choice for $q_{\vtheta}(\va_t,\vu_t|\vs_t,k)$ enables us to solve the sub-optimization w.r.t.~$\vtheta$ by using RL with reward function $r_{\vphi}$. Meanwhile, the choice for $p_{\vomega}(\naction_t|\state_t,\action_t, k)$ incorporates our prior knowledge that the noisy policy tends to Gaussian, which is a reasonable assumption for actual human motor behavior~\citep{vanBeers2004}. Under these model specifications, solving $\max_{\vphi, \vomega, \vpsi} \min_{\vtheta} \mathcal{F}(\vphi, \vomega, \vpsi) - \mathcal{G}(\vphi, \vomega, \vtheta)$ is equivalent to solving $\max_{\vphi, \vomega, \vpsi}\min_{\vtheta} \mathcal{H}(\vphi, \vomega, \vpsi, \vtheta)$, where 
	\begin{align}
	\mathcal{H}(\vphi, \vomega, \vpsi, \vtheta)\!
	&= \!\mathbb{E}_{p_{\mathrm{d}}( \state_{1:T}, \naction_{1:T} | k)p(k) } \! \left[ \textstyle{\sum}_{t=1}^T \mathbb{E}_{q_{\vpsi}(\action_t|\state_t,\naction_t, k)} \! \left[ r_{\vphi}(\state_t, \action_t) \!-\! \frac{1}{2} \|\vu_t - \va_t  \|^2_{\vC^{-1}_{\vomega}(k)} \right] \!+\! H_t(q_{\vpsi}) \right] \notag \\
	&\phantom{=} \!- \mathbb{E}_{\widetilde{q}_{\vtheta}(\vs_{1:T}, \va_{1:T})} \! \left[ \textstyle{\sum}_{t=1}^T {r_{\vphi}(\state_t, \action_t)} \!-\! \log q_{\vtheta}(\va_t|\vs_t)  \right] \!+\! \frac{T}{2} \mathbb{E}_{p(k)} \! \left[ \mathrm{Tr}(\vC_{\vomega}^{-1}(k)\vSigma) \right]. \label{eq:obj_final}
	\end{align}
	Here, $\widetilde{q}_{\vtheta}(\vs_{1:T}, \va_{1:T}) = p_1(\vs_1)\Pi_{t=1}^T \int_{\mathbb{R}} p(\vs_{t+1}| \vs_t, \va_t+\vepsilon_t) \mathcal{N}(\vepsilon_t|0,\vSigma) \mathrm{d}\vepsilon_t q_{\vtheta}(\va_t|\vs_t)$ is a noisy trajectory density induced by a policy  $q_{\vtheta}(\va_t|\vs_t)$, where  $\mathcal{N}(\vepsilon_t|0,\vSigma)$ can be regarded as an approximation of the noisy policy in Figure~\ref{figure:pgm_cirl}. Minimizing $\mathcal{H}$ w.r.t.~$\vtheta$ resembles solving a MaxEnt-RL problem with a reward function $r_{\vphi}(\vs_t,\va_t)$, except that trajectories are collected according to the noisy trajectory density. In other words, this minimization problem can be solved using RL, and $q_{\vtheta}(\va_t|\vs_t)$ can be regarded as an approximation of the optimal policy. The hyper-parameter $\vSigma$ determines the quality of this approximation: smaller value of $\vSigma$ gives a better approximation. Therefore, by choosing a reasonably small value of $\vSigma$, solving the max-min problem yields a reward function $r_{\vphi}(\va_t|\vs_t)$ and a policy $q_{\vtheta}(\va_t|\vs_t)$. This policy imitates the optimal policy, which is the goal of IL.
	
	We note that the model assumption for $p_{\vomega}$ incorporates our prior knowledge about the noisy policy $p(\naction_t|\state_t,\action_t, k)$. Namely, $p_{\vomega}(\naction_t|\state_t,\action_t, k) = \mathcal{N}(\vu_t | \va_t, \vC_{\vomega}(k))$ assumes that the noisy policy tends to Gaussian, where the covariance $\vC_{\vomega}(k)$ gives an estimated expertise of the $k$-th demonstrator: High-expertise demonstrators have small $\vC_{\vomega}(k)$ and vice-versa for low-expertise demonstrators. VILD is not restricted to this choice. Different choices of $p_{\vomega}$ incorporate different prior knowledge. For example, we may use a Laplace distribution to incorporate a prior knowledge about demonstrators who tend to execute outlier actions~\citep{murphy2013}. In such a case, the squared error in $\mathcal{H}$ is simply replaced by the absolute error (see Appendix~\ref{appendix:obj_H}). 
	
	It should be mentioned that $q_{\vpsi}(\va_t|\vs_t,\vu_t,k)$ is a maximum-entropy probability density which maximizes the immediate reward at time $t$ and minimizes the weighted squared error between $\vu_t$ and $\va_t$. The trade-off between the reward and squared-error is determined by the covariance $\vC_{\vomega}(k)$. Specifically, for demonstrators with a small $\vC_{\vomega}(k)$ (i.e., high-expertise demonstrators), the squared error has a large magnitude and $q_{\vpsi}$ tends to minimize the squared error. Meanwhile, for demonstrators with a large value of $\vC_{\vomega}(k)$ (i.e., low-expertise demonstrators), the squared error has a small magnitude and $q_{\vpsi}$ tends to maximize the immediate reward. 
	
	In practice, we include a regularization term $L(\vomega) = T\mathbb{E}_{p(k)}[ \log|\vC^{-1}_{\vomega}(k)| ]/2$, to penalize large covariance. Without this regularization, the covariance can be overly large which makes learning degenerate. We note that $\mathcal{H}$ already includes such a penalty via the trace term: $\mathbb{E}_{p(k)}[ \mathrm{Tr}(\vC_{\vomega}^{-1}(k)\vSigma)]$. However, the strength of this penalty tends to be too small, since we choose $\vSigma$ to be small.

	\subsection{Importance sampling for reward learning}
	\label{section:is}
	\vspace{-1mm}

	To improve the convergence rate of VILD when updating the reward parameter $\vphi$, we use importance sampling (IS). 
	Specifically, by analyzing the gradient $\nabla_{\vphi} \mathcal{H} = \nabla_{\vphi} \{ \mathbb{E}_{p_{\mathrm{d}}( \state_{1:T}, \naction_{1:T}| k)p(k) } [ \Sigma_{t=1}^T \mathbb{E}_{q_{\vpsi}(\action_t|\state_t,\naction_t, k)} [  r_{\vphi}(\state_t, \action_t) ] ] - \mathbb{E}_{\widetilde{q}_{\vtheta}(\vs_{1:T}, \va_{1:T})} [ \Sigma_{t=1}^T r_{\vphi}(\state_t, \action_t) ] \}$, we can see that the reward function is updated to maximize expected cumulative rewards obtained by demonstrators and $q_{\vpsi}$, while minimizing expected cumulative rewards obtained by $q_{\vtheta}$. However, low-quality demonstrations often have low reward values. 
	For this reason, stochastic gradients estimated by these demonstrations tend to be uninformative, which leads to slow convergence and poor data-efficiency.
	
	To avoid estimating such uninformative gradients, we use IS to estimate gradients using high-quality demonstrations which are sampled with high probability. Briefly, IS is a technique for estimating an expectation over a distribution by using samples from a different distribution~\citep{Robert2005}. For VILD, we propose to sample $k$ from a distribution $\tilde{p}(k) = {z_k}/{\Sigma_{k'=1}^{K} z_{k'}}$, where $z_k = \|\mathrm{vec}(\vC^{-1}_{\vomega}(k))\|_1$. This distribution assigns high probabilities to demonstrators with high estimated level of expertise. With this distribution, the estimated gradients tend to be more informative which leads to a faster convergence. To reduce a sampling bias, we use a truncated importance weight: $w(k) = \mathrm{min}({p(k)/\tilde{p}(k)}, 1)$~\citep{Ionides2008}. The distribution $\tilde{p}(k)$ and the importance weight $w(k)$ lead to an IS gradient: $\nabla_{\vphi}\mathcal{H}_{\mathrm{IS}} = \nabla_{\vphi} \lbrace \mathbb{E}_{p_{\mathrm{d}}( \state_{1:T}, \naction_{1:T} | k) \tilde{p}(k) } [ w(k) \Sigma_{t=1}^T \mathbb{E}_{q_{\vpsi}(\action_t|\state_t,\naction_t, k)} [ {r_{\vphi}(\state_t, \action_t)}{} ] ] - \mathbb{E}_{\widetilde{q}_{\vtheta}(\vs_{1:T}, \va_{1:T})} [ \Sigma_{t=1}^T {r_{\vphi}(\state_t, \action_t)} ] \rbrace$. 
	Computing the importance weight requires $p(k)$, which can be estimated accurately since $k$ is a discrete random variable. For simplicity, we assume that $p(k)$ is a uniform distribution. 
	A pseudo-code of VILD with IS is given in Algorithm~\ref{algo:code_main} and more details of our implementation are given in Appendix~\ref{appendix:algo}.
	
	
	\begin{algorithm}[t]
		\caption{VILD: Variational Imitation Learning with Diverse-quality demonstrations}
		\label{algo:code_main}
		\begin{algorithmic}[1]
			\State \textbf{Input: } Diverse-quality demonstrations $\mathcal{D}_{\mathrm{d}} = \{ (\vs_{1:T}, \vu_{1:T}, k)_n \}_{n=1}^N$ and a replay buffer $\mathcal{B} = \varnothing$.
			\While{ Not converge }
			\While { $| \mathcal{B} | < B$ with batch size $B$ }	\Comment{Collect samples from $\widetilde{q}_{\vtheta}(\vs_{1:T},\va_{1:T})$}
			\State Sample $\va_t \sim q_{\vtheta}(\va_t|\vs_t)$ and $\vepsilon_t \sim \mathcal{N}(\vepsilon_t|\boldsymbol{0},\vSigma)$.
			\State Execute  $\va_t + \vepsilon_t$ in environment and observe next state $\vs_{t}' \sim p(\vs_{t}' | \vs_t, \va_t + \vepsilon_t)$.
			\State Include $(\vs_t,\va_t,\vs'_t)$ into the replay buffer $\mathcal{B}$. Set $t \leftarrow t+1$.
			\EndWhile
			\State Update $q_{\vpsi}$ by an estimate of ${\nabla}_{\vpsi} \mathcal{H}(\vphi, \vomega, \vpsi, \vtheta)$.
			\State Update $p_{\vomega}$ by an estimate of ${\nabla}_{\vomega} \mathcal{H}(\vphi, \vomega, \vpsi, \vtheta) + {\nabla}_{\vomega} L(\vomega)$.
			\State Update $r_{\vphi}$ by an estimate of $\nabla_{\vphi} \mathcal{H}_{\mathrm{IS}}(\vphi, \vomega, \vpsi, \vtheta)$.
			\State Update $q_{\vtheta}$ by an RL method (e.g., TRPO or SAC) with reward function $r_{\vphi}$.
			\EndWhile 
		\end{algorithmic}
	\end{algorithm}
	
	\vspace{-2mm}
	\section{Experiments}
	\label{section:experiment}
	\vspace{-2mm}
	
	In this section, we experimentally evaluate the performance of VILD (with and without IS) in Mujoco tasks from OpenAI Gym~\citep{BrockmanEtAl2016}. 
	Performance is evaluated using cumulative ground-truth rewards along trajectories (i.e., higher is better), which is computed using 10 test trajectories generated by learned policies (i.e., $q_{\vtheta}(\va_t|\vs_t)$). We repeat experiments for 5 trials with different random seeds and report the mean and standard error.
	
	\vspace{-3mm}
	\paragraph{Baselines \& data generation.} 
	We compare VILD against GAIL~\citep{HoE16}, AIRL~\citep{FuEtAl2018}, VAIL~\citep{peng2018variational}, MaxEnt-IRL~\citep{ZiebartEtAl2010}, and InfoGAIL~\citep{LiSE17}. These are online IL methods which collect transition samples to learn policies. We use trust-region policy optimization (TRPO)~\citep{SchulmanEtAl2015} to update policies, except for the Humanoid task where we use soft actor-critic (SAC)~\citep{HaarnojaZAL18}. To generate demonstrations from $\pi^\star$ (pre-trained by TRPO) according to Figure~\ref{figure:pgm_cirl}, we use two types of noisy policy $p(\vu_t|\va_t,\vs_t,k)$: Gaussian noisy policy: $\mathcal{N}(\vu_t | \va_t, \sigma^2_{k}\bI)$ and time-signal-dependent (TSD) noisy policy: $\mathcal{N}(\vu_t | \va_t, \diag(\vb_{k}(t) \times \| \va_t \|_1) )$, where $\vb_k(t)$ is sampled from a noise process. We use 10 demonstrators with different $\sigma_k$ and noise processes for $\vb_{k}(t)$.
	Notice that for TSD, the noise variance depends on time and magnitude of actions. This characteristic of TSD has been observed in human motor control~\citep{vanBeers2004}. More details of data generation are given in Appendix~\ref{appendix:setting}.
	
\begin{figure}[!t]
		\centering
		\begin{subfigure}[b]{0.99\linewidth}
			\centering
			\includegraphics[width=0.99\linewidth]{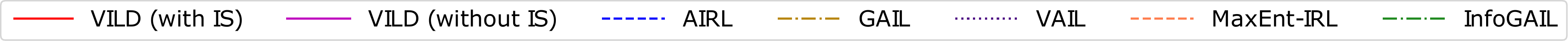}
		\end{subfigure}	
		
		\vspace{1mm}
		
		\centering
		\begin{subfigure}[b]{0.48\linewidth}
			\centering
			\includegraphics[width=0.48\linewidth]{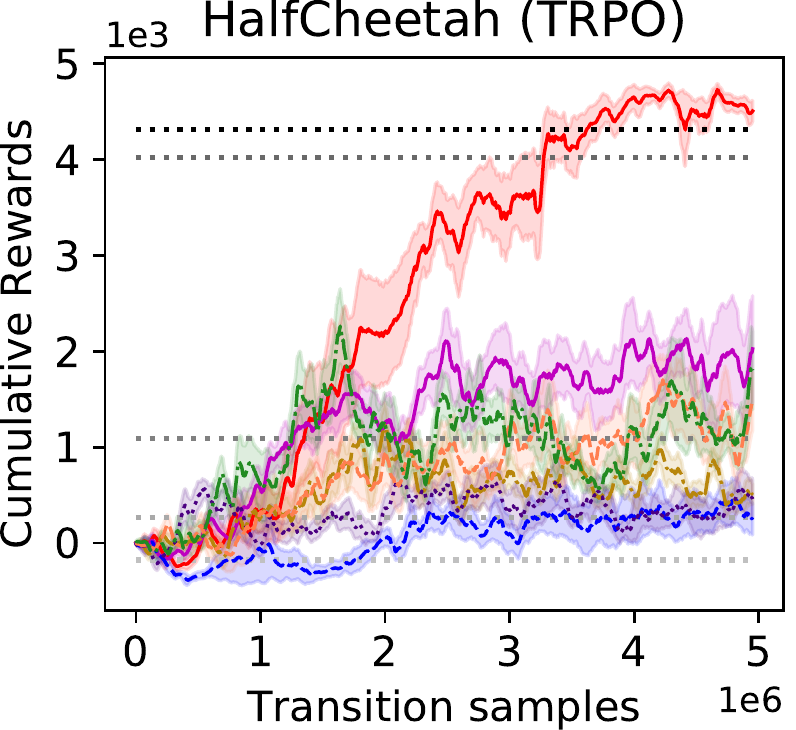}
			\includegraphics[width=0.48\linewidth]{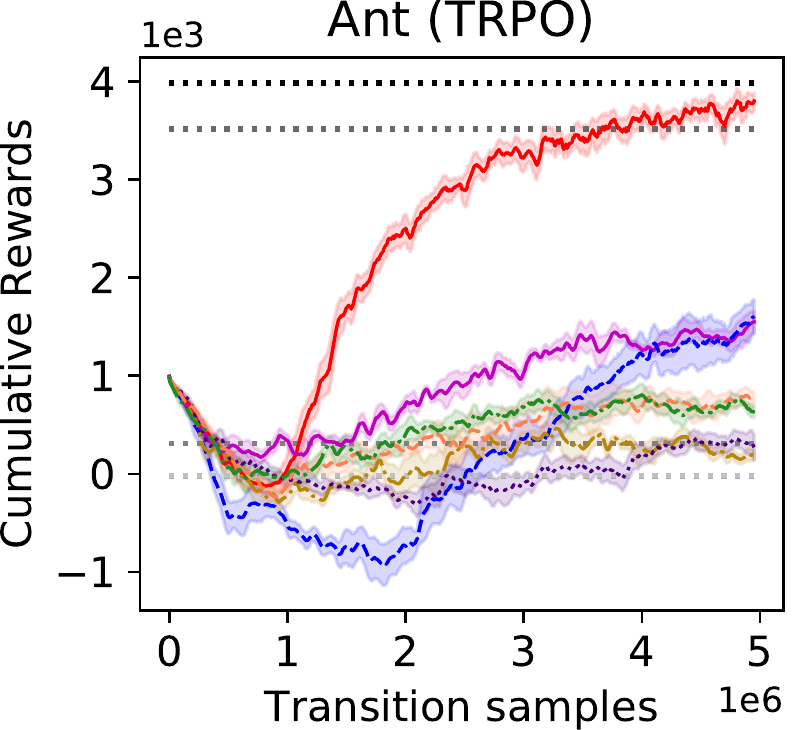}
			\hfill
			\subcaption{Performan when demonstrations are generated using Gaussian noisy policy.}
			\label{figure:exp_trpo_normal}
		\end{subfigure}	
		\hfill
		\begin{subfigure}[b]{0.48\linewidth}
			\centering
			\includegraphics[width=0.48\linewidth]{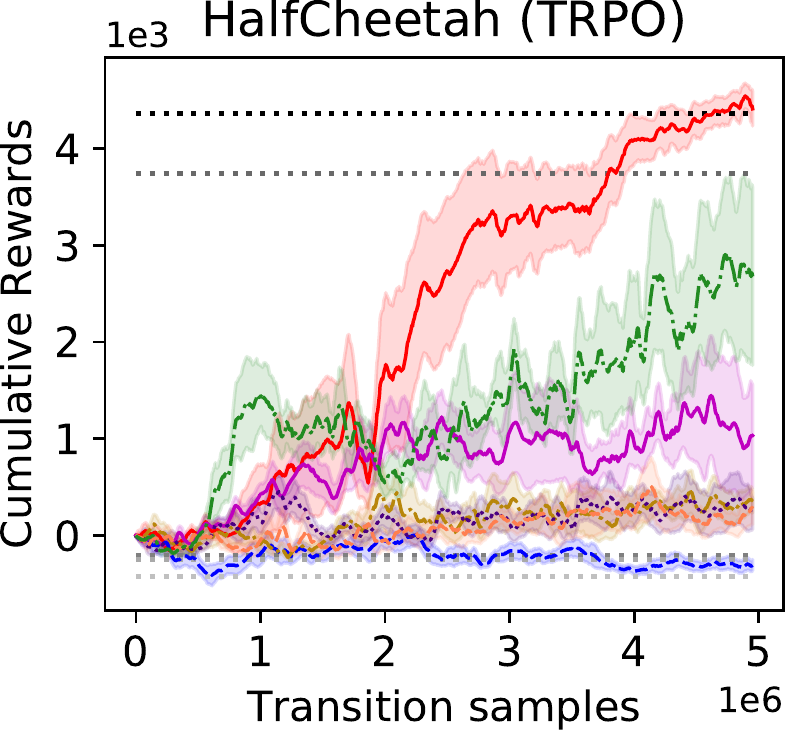}
			\includegraphics[width=0.48\linewidth]{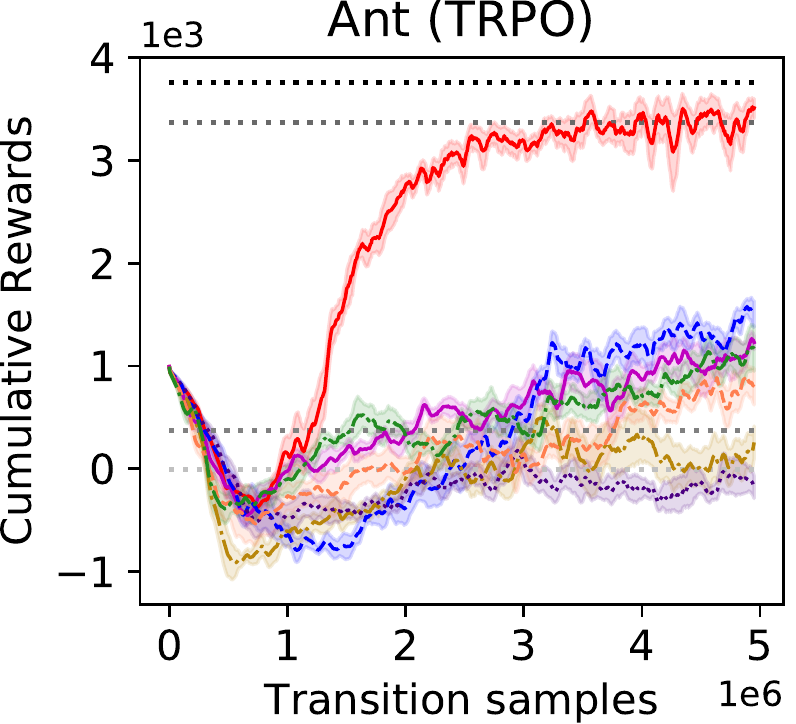}
			\hfill
			\subcaption{Performan when demonstrations are generated using TSD noisy policy.}
			\label{figure:exp_trpo_SDNTN}
		\end{subfigure}	
		\caption{Performance averaged over 5 trials in terms of the mean and standard error. Demonstrations are generated by 10 demonstrators using (a) Gaussian and (b) TSD noisy policies. Horizontal dotted lines indicate performance of $k=1,3,5,7,10$ demonstrators. IS denotes importance sampling.}	
		\label{figure:exp_main}
	\end{figure}

	\vspace{-3mm}
	\paragraph{Results against online IL methods.}
	Figure~\ref{figure:exp_main} shows learning curves of  VILD and existing methods against the number of transition samples in HalfCheetah and Ant\footnote{Learning curves of other tasks are given in Appendix~\ref{appendix:more_exp}.}, whereas Table~\ref{table:performance} reports the performance achieved in the last 100 update iterations. We can see that VILD with IS outperforms existing methods in terms of both data-efficiency and final performance, i.e., VILD with IS learns better policies using less numbers of transition samples. VILD without IS tends to outperform existing methods in terms of the final performance. However, it is less data-efficient when compared to VILD with IS, except on Humanoid with the Gaussian noisy policy, where VILD without IS performs better than VILD with IS in terms of the final performance. 
	We conjecture that this is because IS slightly biases gradient estimation, which may have a negative effect on the performance.
	Nonetheless, the overall good performance of VILD with IS suggests that it is an effective method to handle diverse-quality demonstrations. 
	
	On the contrary, existing methods perform poorly overall. We found that InfoGAIL, which learns a context-dependent policy, can achieve good performance when the policy is conditioned on specific contexts. However, its performance is quite poor on average when using contexts sampled from a (uniform) prior distribution. These results supports our conjecture that existing methods are not suitable for diverse-quality demonstrations when the level of demonstrators' expertise in unknown.
	
	It can be seen that VILD without IS performs better for the Gaussian noisy policy when compared to the TSD noisy policy. This is because the model of VILD is correctly specified for the Gaussian noisy policy, but the model is incorrectly specified for the TSD noisy policy; misspecified model indeed leads to the reduction in performance. Nonetheless, VILD with IS still perform well for both types of noisy policy. This is perhaps because negative effects of a misspecified model is not too severe for learning expertise parameters, which are required to compute $\widetilde{p}(k)$. 
	
	We also conduct the following evaluations. Due to space limitation, figures are given in Appendix~\ref{appendix:more_exp}.
	
	\vspace{-3mm}
	\paragraph{Results against offline IL methods.}
	We compare VILD against offline IL methods based on supervised learning, namely behavior cloning (BC)~\citep{Pomerleau88}, Co-Teaching which is based on a noisy label learning method~\citep{HanYYNXHTS18}, and BC from diverse-quality demonstrations (BC-D) which optimizes the naive model described in Section~\ref{section:model}. 
	Results in Figure~\ref{figure:exp_app_offline} show that these methods perform worse than VILD overall; BC performs the worst since it severely suffers from both the compounding error and low-quality demonstrations. BC-D and Co-teaching are quite robust against low-quality demonstrations, but they perform poorly due to the issue of compounding error.

	\vspace{-3mm}
	\paragraph{Accuracy of estimated expertise parameter.}
	To evaluate accuracy of estimated expertise parameter, we compare the ground-truth value of $\sigma_k$ under the Gaussian noisy policy against the learned covariance $\vC_{\vomega}(k)$. Figure~\ref{figure:exp_app_online_expertise} shows that VILD learns an accurate ranking of demonstrators' expertise. The values of these parameters are also quite accurate compared to the ground-truth, except for demonstrators with low-levels of expertise.
	A reason for this phenomena is that low-quality demonstrations are highly dissimilar, which makes learning the expertise more challenging.

	\begin{table}
		\caption{
			Performance in the last 100 iterations in terms of the mean and standard error of cumulative rewards (higher is better). (G) denotes Gaussian noisy policy and (TSD) denotes time-signal-dependent noisy policy. Boldfaces indicate best and comparable methods according to t-test with p-value 0.01. The performance of VAIL is similar to that of GAIL and is omitted.}
		\label{table:performance}
		\small
		\begin{tabular}{l || r | r | r | r | r | r }
			\hline
			Task & VILD (IS) & VILD (w/o IS) & AIRL & GAIL & MaxEnt-IRL & InfoGAIL \\ 
			\hline \hline
			HalfCheetah (G)  & \textbf{4559 (43)} & 1848 (429) & 341 (177) & 551 (23) & 1192 (245) & 1244 (210) \\
			HalfCheetah (TSD) & \textbf{4394 (136)} & 1159 (594) & -304 (51) & 318 (134) & 177 (132) & \textbf{2664 (779)} \\
			\hline
			Ant (G) & \textbf{3719 (65)} & 1426 (81) & 1417 (184) & 209 (30) & 731 (93) & 675 (36) \\
			Ant (TSD) & \textbf{3396 (64)} & 1072 (134) & 1357 (59) & 97 (161) & 775 (135) & 1076 (140) \\
			\hline
			Walker2d (G) & \textbf{3470 (300)} & 2132 (64) & 1534 (99) & 1410 (115) & 1795 (172) & 1668 (82) \\
			Walker2d (TSD) & \textbf{3115 (130)} & 1244 (132) & 578 (47) & 834 (84) & 752 (112) & 1041 (36) \\
			\hline
			Humanoid (G) & \textbf{3781 (557)} & \textbf{4840 (56)} & 4274 (93) & 284 (24) & \textbf{3038 (731)} & \textbf{4047 (653)} \\
			Humanoid (TSD) & \textbf{4600 (97)} & \textbf{3610 (448)} & \textbf{4212 (121)} & 203 (31) & \textbf{4132 (651)} & \textbf{3962 (635)} \\
			

		\end{tabular}
		
	\end{table}

	\vspace{-2mm}
	\section{Conclusion and Future Work}
	\vspace{-1mm}
	
	In this paper, we explored a practical setting of IL where demonstrations have diverse-quality. We showed the deficiency of existing methods, and proposed a robust method called VILD which learns both the reward function and the level of demonstrators' expertise by using the variational approach. Empirical results demonstrated that our work enables scalable and data-efficient IL under this practical setting. In future, we will explore other approaches to efficiently estimate parameters of the proposed model except the variational approach.

	\bibliography{my_lib.bib}
	\bibliographystyle{plainnat}

\appendix 	

\section{Derivations}
\label{appendix:proof}

This section derives the lower-bounds of $f(\vphi,\vomega)$ and $g(\vphi,\vomega)$ presented in the paper. 
We also derive the objective function $\mathcal{H}(\vphi,\vomega,\vpsi,\vtheta)$ of VILD.

\subsection{Lower-bound of $f$}

Let $l_{\vphi,\vomega}(\vs_t,\va_t,\vu_t,k) = r_{\vphi}(\vs_t,\va_t) + \log p_{\vomega}(\vu_t|\vs_t,\va_t,k)$, we have that $f(\vphi, \vomega) = \mathbb{E}_{p_{\mathrm{d}}(\state_{1:T}, \naction_{1:T}| k)p(k)} \left[ \sum_{t=1}^T f_t(\vphi, \vomega) \right]$, 
where $f_t(\vphi, \vomega) = \log \int_{\mathcal{A}} \exp\left( l_{\vphi,\vomega}(\vs_t,\va_t,\vu_t,k) \right) \daction_t$. By using a variational distribution $q_{\vpsi}(\va_t|\vs_t,\vu_t,\id)$ with parameter $\vpsi$, we can bound $f_t(\vphi, \vomega)$ from below by using the Jensen inequality as follows:
\begin{align}
f_t(\vphi, \vomega)
&= \log \left( \int_{\mathcal{A}} \exp\left( l_{\vphi,\vomega}(\vs_t,\va_t,\vu_t,k) \right) \frac{q_{\vpsi}(\va_t|\vs_t,\vu_t,\id)}{q_{\vpsi}(\va_t|\vs_t,\vu_t,\id)}  \daction_t \right)  \notag \\
&\geq \int_{\mathcal{A}} q_{\vpsi}(\va_t|\vs_t,\vu_t,\id) \log \left( \exp\left( l_{\vphi,\vomega}(\vs_t,\va_t,\vu_t,k) \right) \frac{1}{q_{\vpsi}(\va_t|\vs_t,\vu_t,\id)} \right) \daction_t \notag \\
&= \mathbb{E}_{q_{\vpsi}(\va_t|\vs_t,\vu_t,\id)}\left[ l_{\vphi,\vomega}(\vs_t,\va_t,\vu_t,k) - \log q_{\vpsi}(\va_t|\vs_t,\vu_t,\id) \right] \notag \\
&= \mathcal{F}_t(\vphi,\vomega,\vpsi). \label{eq:f_bound_summand}
\end{align}
Then, by using the linearity of expectation, we obtain the lower-bound of $f(\vphi,\vomega)$ as follows:
\begin{align}
f(\vphi, \vomega)
&\geq \mathbb{E}_{p_{\mathrm{d}}(\state_{1:T}, \naction_{1:T}| k)p(k)} \left[ \textstyle{\sum}_{t=1}^T \mathcal{F}_t(\vphi,\vomega,\vpsi) \right] \notag \\
&= \mathbb{E}_{p_{\mathrm{d}}(\state_{1:T}, \naction_{1:T}| k)p(k)} \left[ \textstyle{\sum}_{t=1}^T \mathbb{E}_{q_{\vpsi}(\va_t|\vs_t,\vu_t,\id)}\left[ l_{\vphi,\vomega}(\vs_t,\va_t,\vu_t,k) - \log q_{\vpsi}(\va_t|\vs_t,\vu_t,\id) \right] \right] \notag \\
&= \mathcal{F}(\vphi,\vomega,\vpsi).
\end{align}

To verify that $f(\vphi,\vomega) = \max_{\vpsi} \mathcal{F}(\vphi,\vomega,\vpsi)$, we maximize $\mathcal{F}_t(\vphi,\vomega,\vpsi)$ w.r.t.~$q_{\vpsi}$ under the constraint that $q_{\vpsi}$ is a valid probability density, i.e., $q_{\vpsi}(\va_t|\vs_t,\vu_t,\id) > 0$ and $\int_{\mathcal{A}} q_{\vpsi}(\va_t|\vs_t,\vu_t,\id) \da_t = 1$.
By setting the derivative of $\mathcal{F}_t(\vphi,\vomega,\vpsi)$ w.r.t.~$q_{\vpsi}$ to zero, we obtain
\begin{align}
q_{\vpsi}(\va_t|\vs_t,\vu_t,\id) &= \exp \left( l_{\vphi,\vomega}(\vs_t,\va_t,\vu_t,k) - 1 \right) \notag \\
&= \frac{ \exp \left( l_{\vphi,\vomega}(\vs_t,\va_t,\vu_t,k) \right)}{\int_{\mathcal{A}} \exp \left( l_{\vphi,\vomega}(\vs_t,\va_t,\vu_t,k) \right) \da_t}, \notag
\end{align}
where the last line follows from the constraint $\int_{\mathcal{A}} q_{\vpsi}(\va_t|\vs_t,\vu_t,\id) \da_t = 1$.
To show that this is indeed the maximizer, we substitute $q_{\vpsi^\star}(\va_t|\vs_t,\vu_t,\id) = \frac{ \exp \left( l(\vs_t,\va_t,\vu_t,k) \right)}{\int_{\mathcal{A}} \exp \left( l(\vs_t,\va_t,\vu_t,k) \right) \da_t}$ into $\mathcal{F}_t(\vphi,\vomega,\vpsi)$:
\begin{align}
\mathcal{F}_t(\vphi,\vomega,\vpsi^\star) 
&= \mathbb{E}_{q^\star_{\vpsi}(\va_t|\vs_t,\vu_t,\id)} \left[ l_{\vphi,\vomega}(\vs_t,\va_t,\vu_t,k) - \log q_{\vpsi^\star}(\va_t|\vs_t,\vu_t,\id) \right] \notag \\
&= \log \left( \int_{\mathcal{A}} \exp \left( l_{\vphi,\vomega}(\vs_t,\va_t,\vu_t,k) \right) \da_t \right). \notag
\end{align}
This equality verifies that $f_t(\vphi,\vomega) = \max_{\vpsi} \mathcal{F}_t(\vphi, \vomega,\vpsi)$.
Finally, by using the linearity of expectation, we have that $f(\vphi,\vomega) = \max_{\vpsi} \mathcal{F}(\vphi, \vomega,\vpsi)$.

\subsection{Lower-bound of $g$}

Next, we derive the lower-bound of $g(\vphi,\vomega)$ presented in the paper. 
We first derive a trivial lower-bound using a general variational distribution over trajectories and reveal its issues.
Then, we derive a lower-bound stated presented in the paper by using a structured variational distribution.
Recall that the function $g(\vphi,\vomega) = \log Z_{\vphi,\vomega}$ is 
\begin{align}
g(\vphi, \vomega)
&= \log \left( \sum_{k=1}^{K} p(k) \!\!\!\!\! \idotsint\displaylimits_{~~~(\mathcal{S} \times \mathcal{A} \times \mathcal{A})^T}
\!\!\! p_1(\state_1)
\prod_{t=1}^T p(\state_{t+1}|\state_t, \naction_t) \exp \left( l(\vs_t,\va_t,\vu_t,k) \right) 
\dstate_{1:T} \dnaction_{1:T} \da_{1:T}\right) \notag.
\end{align}

\paragraph{Lower-bound via a variational distribution}
A lower-bound of $g$ can be obtained by using a variational distribution $\widebar{q}_{\vbeta}(\vs_{1:T}, \vu_{1:T}, \va_{1:T}, k)$ with parameter $\vbeta$. We note that this variational distribution allows any dependency between the random variables $\vs_{1:T}$, $\vu_{1:T}$, $\va_{1:T}$, and $k$. By using this distribution, we have a lower-bound
\begin{align}
g(\vphi,\vomega)
&=  \log \Bigg( \sum_{k=1}^{K} p(k) \!\!\!\!\! \idotsint\displaylimits_{~~~(\mathcal{S} \times \mathcal{A} \times \mathcal{A})^T}
\!\!\! p_1(\state_1) \prod_{t=1}^T p(\state_{t+1}|\state_t, \naction_t) \exp \left( l_{\vphi,\vomega}(\vs_t,\va_t,\vu_t,k) \right) \notag \\
&\phantom{=} \quad \quad \times  
\frac{\widebar{q}_{\vbeta}(\vs_{1:T}, \vu_{1:T}, \va_{1:T}, k)}{\widebar{q}_{\vbeta}(\vs_{1:T}, \vu_{1:T}, \va_{1:T}, k)}
\dstate_{1:T} \dnaction_{1:T} \da_{1:T} \Bigg) \notag \\
&\geq \mathbb{E}_{\widebar{q}_{\vbeta}(\vs_{1:T}, \vu_{1:T}, \va_{1:T}, k)} \Bigg[ 
\log p(k) p_1(\state_1) + \sum_{t=1}^T \left\lbrace \log p(\state_{t+1}|\state_t, \naction_t) + l_{\vphi,\vomega}(\vs_t,\va_t,\vu_t,k) \right\rbrace \notag \\
&\phantom{=} \quad\quad - \log \widebar{q}_{\vbeta}(\vs_{1:T}, \vu_{1:T}, \va_{1:T}, k) \Bigg] \notag \\
&= \widebar{\mathcal{G}}(\vphi,\vomega,\vbeta).
\end{align}
The main issue of using this lower-bound is that, $\widebar{\mathcal{G}}(\vphi,\vomega,\vbeta)$ can be computed or approximated only when we have an access to  the transition probability $p(\state_{t+1}|\state_t, \naction_t)$. In many practical tasks, the transition probability is unknown and needs to be approximated. However, approximating the transition probability for large state and action spaces is known to be highly challenging~\citep{SzitaS10}. For these reasons, this lower-bound is not suitable for our method.

\paragraph{Lower-bound via a structured variational distribution}
\label{app:bound_g_structure}

To avoid the above issue, we use the structure variational approach~\citep{HoffmanB15}, where the key idea is to pre-define conditional dependenc to ease computation. Specifically, we use  a variational distribution $q_{\vtheta}(\action_t, \vu_t| \state_t, k)$ with parameter $\vtheta$ and define dependencies between states according to the transition probability of MDPs. With this variational distribution, we lower-bound $g$ as follows:
\begin{align}
g(\vphi,\vomega) 
&=  \log \Bigg( \sum_{k=1}^{K} p(k) \!\!\!\!\! \idotsint\displaylimits_{~~~(\mathcal{S} \times \mathcal{A} \times \mathcal{A})^T}
\!\!\! p_1(\state_1) \prod_{t=1}^T p(\state_{t+1}|\state_t, \naction_t) \exp \left( l_{\vphi,\vomega}(\vs_t,\va_t,\vu_t,k) \right) \notag \\
&\phantom{=} \quad\quad \times \frac{ q_{\vtheta}(\action_t, \vu_t| \state_t, k) }{ q_{\vtheta}(\action_t, \vu_t| \state_t, k) }
\dstate_{1:T} \dnaction_{1:T} \da_{1:T} \Bigg) \notag \\
&\geq \mathbb{E}_{\widetilde{q}_{\vtheta}(\vs_{1:T},\vu_{1:T},\va_{1:T}, k)} \left[ \sum_{t=1}^T l_{\vphi,\vomega}(\vs_t,\va_t,\vu_t,k) - \log q_{\vtheta}(\action_t, \vu_t| \state_t, k) \right] \notag \\
&= \mathcal{G}(\vphi,\vomega,\vtheta),
\end{align}
where $\widetilde{q}_{\vtheta}(\vs_{1:T},\vu_{1:T},\va_{1:T}, k) = p(k) p_1(\state_1) \Pi_{t=1}^T p(\state_{t+1}|\state_t, \naction_t) q_{\vtheta}(\action_t, \vu_t| \state_t, k)$. The optimal variational distribution $q_{\vtheta^\star}(\va_t,\vu_t|\vs_t,k)$ can be founded by maximizing $\mathcal{G}(\vphi,\vomega,\vtheta)$ w.r.t.~$q_{\vtheta}$. 
Solving this maximization problem is identical to solving a maximum entropy RL (MaxEnt-RL) problem~\citep{ZiebartEtAl2010} for an MDP defined by a tuple ${\mathcal{M}} = (\mathcal{S} \times \mathbb{N}_+, \mathcal{A} \times \mathcal{A}, p(\vs', | \vs, \vu) \mathbb{I}_{\id = \id'}, p_1(\vs_1)p(\id_1), l_{\vphi,\vomega}(\vs,\va,\vu,k) )$. Specifically, this MDP is defined with a state variable $(\vs_t, \id_t) \in \mathcal{S} \times \mathbb{N}$, an action variable $(\va_t, \vu_t) \in \mathcal{A} \times \mathcal{A}$, a transition probability density $p(\vs_{t+1}, | \vs_{t}, \vu_t) \mathbb{I}_{\id_t = \id_{t+1}}$, an initial state density $p_1(\vs_1)p(\id_1)$, and a reward function $l_{\vphi,\vomega}(\vs_t,\va_t,\vu_t,k)$. Here, $\mathbb{I}_{a = b}$ is the indicator function which equals to $1$ if $a = b$ and $0$ otherwise. 
By adopting the optimality results of MaxEnt-RL~\citep{ZiebartEtAl2010,Levine2018}, we have $g(\vphi,\vomega) = \max_{\vtheta}\mathcal{G}(\vphi,\vomega,\vtheta)$, where the optimal variational distribution is
\begin{align}
q_{\vtheta^\star}(\action_t, \vu_t|\vs_t,k) = \exp( Q(\vs_t,k,\va_t,\vu_t) - V(\vs_t,k)).
\end{align}
The functions $Q$ and $V$ are soft-value functions defined as 
\begin{align}
Q(\vs_t,k,\va_t,\vu_t) &= l_{\vphi,\vomega}(\vs_t,\va_t,\vu_t,k) + \mathbb{E}_{p(\state_{t+1}|\state_t,\naction_t) } \left[ V( \vs_{t+1}, k ) \right], \\
V(\vs_t,k) &=  \log \iint_{\mathcal{A} \times \mathcal{A}} \exp\left( Q(\vs_t,k,\va_t,\vu_t)  \right) \daction_t\dnaction_t.
\end{align}

\subsection{Objective function $\mathcal{H}$ of VILD}
\label{appendix:obj_H}

This section derives the objective function $\mathcal{H}(\vphi,\vomega,\vpsi,\vtheta)$ from $\mathcal{F}(\vphi,\vomega,\vpsi) - \mathcal{G}(\vphi,\vomega,\vtheta)$. Specfically, we substitute the models $p_{\vomega}(\naction_t|\state_t,\action_t, \id) = \mathcal{N}(\vu_t | \va_t, \vC_{\vomega}(\id))$ and $q_{\vtheta}(\va_t,\vu_t|\vs_t,k) =  q_{\vtheta}(\va_t|\vs_t) \mathcal{N}(\vu_t | \va_t, \vSigma)$. We also give an example when using a Laplace distribution for $p_{\vomega}(\naction_t|\state_t,\action_t, \id)$ instead of the Gaussian distribution.


First, we substitute $q_{\vtheta}(\va_t,\vu_t|\vs_t,k) =  q_{\vtheta}(\va_t|\vs_t) \mathcal{N}(\vu_t | \va_t, \vSigma)$ into $\mathcal{G}$:
\begin{align*}
\mathcal{G}(\vphi, \vomega, {\vtheta}) 
&= \mathbb{E}_{\widetilde{q}_{\vtheta}(\state_{1:T}, \naction_{1:T}, \action_{1:T} , \id)}	\left[ 
\sum_{t=1}^T l_{\vphi,\vomega}(\vs_t,\va_t,\vu_t,k) - \log \mathcal{N}(\vu_t | \va_t, \vSigma) - \log q_{\vtheta}(\action_t | \state_t)  \right] \\
&= \mathbb{E}_{\widetilde{q}_{\vtheta}(\state_{1:T}, \naction_{1:T}, \action_{1:T}, \id)}	\left[ 
\sum_{t=1}^T l_{\vphi,\vomega}(\vs_t,\va_t,\vu_t,k) + \frac{1}{2} \|\vu_t - \va_t  \|^2_{\vSigma^{-1}} - \log q_{\vtheta}(\action_t | \state_t) \right] + \mathrm{c_1},
\end{align*}
where $c_1$ is a constant corresponding to the log-normalization term of the Gaussian distribution. 
Next, by using the re-parameterization trick, we rewrite $\widetilde{q}_{\vtheta}(\state_{1:T}, \naction_{1:T}, \action_{1:T} , \id)$ as
\begin{align}
\widetilde{q}_{\vtheta}(\state_{1:T}, \naction_{1:T}, \action_{1:T}, \id) = p(\id) p_1(\vs_1) \prod_{t=1}^T p_1(\vs_{t+1}|\state_t, \va_t + \vSigma^{1/2}\vepsilon_t) \mathcal{N}(\vepsilon_t | 0, \bI) q_{\vtheta}(\va_t|\vs_t), \notag
\end{align}
where we use $\vu_t = \va_t + \vSigma^{1/2}\vepsilon_t$ with $\vepsilon_t \sim \mathcal{N}(\vepsilon_t|0,\bI)$.
With this, the expectation of $\Sigma_{t=1}^T\|\vu_t - \va_t  \|^2_{\vSigma^{-1}}$ over $\widetilde{q}_{\vtheta}(\state_{1:T}, \naction_{1:T}, \action_{1:T}, \id)$ can be written as
\begin{align*}
\mathbb{E}_{\widetilde{q}_{\vtheta}(\state_{1:T}, \naction_{1:T}, \action_{1:T}, \id)} \left[ \sum_{t=1}^T \|\vu_t - \va_t  \|^2_{\vSigma^{-1}} \right] 
&=
\mathbb{E}_{\widetilde{q}_{\vtheta}(\state_{1:T}, \naction_{1:T}, \action_{1:T}, \id)} \left[ \sum_{t=1}^T \|\va_t + \vSigma^{1/2}\vepsilon_t - \va_t  \|^2_{\vSigma^{-1}}\right] \\
&=
\mathbb{E}_{\widetilde{q}_{\vtheta}(\state_{1:T}, \naction_{1:T}, \action_{1:T}, \id)} \left[ \sum_{t=1}^T \| \vSigma^{1/2}\vepsilon_t \|^2_{\vSigma^{-1}}\right] \\
&= T\ad,
\end{align*}
which is a constant. Then, the quantity $\mathcal{G}$ can be expressed as
\begin{align*}
\mathcal{G}(\vphi, \vomega, {\vtheta}) 
&= \mathbb{E}_{\widetilde{q}_{\vtheta}(\state_{1:T}, \naction_{1:T}, \action_{1:T}, \id)}	\left[ 
\sum_{t=1}^T l_{\vphi,\vomega}(\vs_t,\va_t,\vu_t,k) - \log q_{\vtheta}(\action_t | \state_t) \right] + c_1 + T{\ad}.
\end{align*}
By ignoring the constant, the optimization problem $\max_{\vphi,\vomega,\vpsi}\min_{\vtheta} \mathcal{F}(\vphi,\vomega,\vpsi) - \mathcal{G}(\vphi,\vomega,\vtheta)$ is equivalent to 
\begin{align}
\max_{\vphi,\vomega,\vpsi}\min_{\vtheta}&~\mathbb{E}_{p_{\mathrm{d}}(\state_{1:T}, \naction_{1:T}, k)} \left[ \sum_{t=1}^T \mathbb{E}_{q_{\vpsi}(\action_t|\state_t,\naction_t, k)} \left[ l_{\vphi,\vomega}(\vs_t,\va_t,\vu_t,k) - \log q_{\vpsi}(\va_t|\vs_t,\vu_t,k) \right] \right] \notag \\
&\phantom{=} - \mathbb{E}_{\widetilde{q}_{\vtheta}(\vs_{1:T}, \vu_{1:T}, \va_{1:T}, k)} \left[ \sum_{t=1}^T l_{\vphi,\vomega}(\vs_t,\va_t,\vu_t,k) - \log q_{\vtheta}(\va_t|\vs_t)  \right].	\label{eq:app_F_G_q}
\end{align}

Our next step is to substitute $p_{\vomega}(\vu_t|\vs_t,\va_t,k)$ by our choice of model. 
First, let us consider a Gaussian distribution $p_{\vomega}(\naction_t|\state_t,\action_t, \id) = \mathcal{N}(\vu_t | \va_t, \vC_{\vomega}(\vs_t,\id))$, where the covariance depends on state.
With this model, the second term in Eq.~\eqref{eq:app_F_G_q} is given by 
\begin{align*}
&\mathbb{E}_{\widetilde{q}_{\vtheta}(\vs_{1:T}, \vu_{1:T}, \va_{1:T}, k)} \left[ \sum_{t=1}^T l_{\vphi,\vomega}(\vs_t,\va_t,\vu_t,k) - \log q_{\vtheta}(\va_t|\vs_t)  \right] \notag \\
&= \mathbb{E}_{\widetilde{q}_{\vtheta}(\state_{1:T}, \naction_{1:T}, \action_{1:T}, \id)}	\left[ 
\sum_{t=1}^T {r_{\vphi}(\state_t, \action_t)} + \log \mathcal{N}(\vu_t | \va_t, \vC_{\vomega}(\vs_t,\id)) - \log q_{\vtheta}(\action_t | \state_t) \right] \\
&= \mathbb{E}_{\widetilde{q}_{\vtheta}(\state_{1:T}, \naction_{1:T}, \action_{1:T}, \id)}	\left[ 
\sum_{t=1}^T {r_{\vphi}(\state_t, \action_t)} - \frac{1}{2} \|\vu_t - \va_t  \|^2_{\vC^{-1}_{\vomega}(\vs_t, \id)} - \frac{1}{2} \log | \vC_{\vomega}(\vs_t, \id) | - \log q_{\vtheta}(\action_t | \state_t) \right]  + c_2,
\end{align*}
where $c_2 = -\frac{\ad}{2} \log {2\pi}$ is a constant.
By using the reparameterization trick, we write the expectation of $\Sigma_{t=1}^T \|\vu_t - \va_t  \|^2_{\vC^{-1}_{\vomega}(\vs_t, \id)}$ as follows:
\begin{align*}
\mathbb{E}_{\widetilde{q}_{\vtheta}(\state_{1:T}, \naction_{1:T}, \action_{1:T}, \id)} \left[ \sum_{t=1}^T \|\vu_t - \va_t  \|^2_{\vC^{-1}_{\vomega}(\vs_t, \id)} \right] 
&=
\mathbb{E}_{\widetilde{q}_{\vtheta}(\state_{1:T}, \naction_{1:T}, \action_{1:T}, \id)} \left[ \sum_{t=1}^T \|\va_t + \vSigma^{1/2}\vepsilon_t - \va_t  \|^2_{\vC^{-1}_{\vomega}(\vs_t, \id)}\right] \\
&=
\mathbb{E}_{\widetilde{q}_{\vtheta}(\state_{1:T}, \naction_{1:T}, \action_{1:T}, \id)} \left[ \sum_{t=1}^T \| \vSigma^{1/2}\vepsilon_t \|^2_{\vC^{-1}_{\vomega}(\vs_t, \id)}\right].
\end{align*}
Using this equality, the second term in Eq.~\eqref{eq:app_F_G_q} is given by
\begin{align}
\mathbb{E}_{\widetilde{q}_{\vtheta}(\state_{1:T}, \naction_{1:T}, \action_{1:T}, \id)}	\left[ 
\sum_{t=1}^T {r_{\vphi}(\state_t, \action_t)} - \log q_{\vtheta}(\action_t | \state_t) 
- \frac{1}{2} \left( \| \vSigma^{1/2}\vepsilon_t \|^2_{\vC^{-1}_{\vomega}(\vs_t, \id)} + \log | \vC_{\vomega}(\vs_t, \id) | \right) \right].
\label{eq:app_max_cov_s}
\end{align}
Maximizing this quantity w.r.t.~$\vtheta$ has an implication as follows: $q_{\vtheta}(\va_t|\vs_t)$ is maximum entropy policy which maximizes expected cumulative rewards while avoiding states that are difficult for demonstrators. 
Specifically, a large value of $\mathbb{E}_{p(\id)} \left[ \log | \vC_{\vomega}(\vs_t, \id) | \right]$ indicates that demonstrators have a low level of expertise for state $\vs_t$ on average, given by our estimated covariance. In other words, this state is difficult to accurately execute optimal actions for all demonstrators on averages. Since the policy $q_{\vtheta}(\va_t|\vs_t)$ should minimize $\mathbb{E}_{p(\id)} \left[ \log | \vC_{\vomega}(\vs_t, \id) | \right]$, the policy should avoid states that are difficult for demonstrators. We expect that this property may improve exploration-exploitation trade-off. Still, such a property is not in the scope of this paper, and we leave it for future work.

In this paper, we assume that the covariance does not depend on state: $\vC_{\vomega}(\vs_t, \id) = \vC_{\vomega}(\id)$. 
This model enables us to simplify Eq.~\eqref{eq:app_max_cov_s} as follows:
\begin{align*}
&\mathbb{E}_{\widetilde{q}_{\vtheta}(\state_{1:T}, \naction_{1:T}, \action_{1:T}, \id)}	\left[ 
\sum_{t=1}^T {r_{\vphi}(\state_t, \action_t)} - \log q_{\vtheta}(\action_t | \state_t) 
- \frac{1}{2} \left( \| \vSigma^{1/2}\vepsilon_t \|^2_{\vC^{-1}_{\vomega}(\id)} + \log | \vC_{\vomega}(\id) | \right) \right] \notag \\
&= \mathbb{E}_{\widetilde{q}_{\vtheta}(\state_{1:T}, \naction_{1:T}, \action_{1:T}, \id)}	\left[ 
\sum_{t=1}^T {r_{\vphi}(\state_t, \action_t)} - \log q_{\vtheta}(\action_t | \state_t) \right] 
- \frac{T}{2} \mathbb{E}_{p(\id) \mathcal{N}(\vepsilon | 0, \bI)} \left[ \| \vSigma^{1/2}\vepsilon \|^2_{\vC^{-1}_{\vomega}(\id)} + \log | \vC_{\vomega}(\id) | \right] \notag \\
&= \mathbb{E}_{\widetilde{q}_{\vtheta}(\vs_{1:T}, \va_{1:T}) }	\left[ 
\sum_{t=1}^T {r_{\vphi}(\state_t, \action_t)} - \log q_{\vtheta}(\action_t | \state_t) \right] 
- \frac{T}{2} \mathbb{E}_{p(\id) } \left[ \mathrm{Tr}(\vC_{\vomega}^{-1}(\id)\vSigma) + \log | \vC_{\vomega}(\id) | \right],
\end{align*}
where $\widetilde{q}_{\vtheta}(\vs_{1:T}, \va_{1:T}) = p_1(\vs_1)\prod_{t=1}^T \int p(\vs_{t+1}| \vs_t, \va_t+\vepsilon_t) \mathcal{N}(\vepsilon_t|0,\vSigma) \mathrm{d}\vepsilon_t q_{\vtheta}(\va_t|\vs_t)$.
The last line follows from the quadratic form identity: $\mathbb{E}_{\mathcal{N}(\vepsilon_t | 0, \bI)} \left[ \| \vSigma^{1/2}\vepsilon_t \|^2_{\vC^{-1}_{\vomega}(\id)} \right] = \mathrm{Tr}(\vC_{\vomega}^{-1}(\id)\vSigma)$. 
Next, we substitute $p_{\vomega}(\naction_t|\state_t,\action_t, \id) = \mathcal{N}(\vu_t | \va_t, \vC_{\vomega}(\id))$ into the first term of Eq.~\eqref{eq:app_F_G_q}.
\begin{align}
&\mathbb{E}_{p_{\mathrm{d}}(\state_{1:T}, \naction_{1:T}, \id)} \left[ \sum_{t=1}^T \mathbb{E}_{q_{\vpsi}(\action_t|\state_t,\naction_t, \id)} \left[ 
l_{\vphi,\vomega}(\vs_t,\va_t,\vu_t,k) - \log q_{\vpsi}(\action_t|\state_t,\naction_t, \id) \right] \right] \notag \\
&= \mathbb{E}_{p_{\mathrm{d}}(\state_{1:T}, \naction_{1:T}, \id)} \Bigg[ \sum_{t=1}^T \mathbb{E}_{q_{\vpsi}(\action_t|\state_t,\naction_t, \id)} \Big[ 
{r_{\vphi}(\state_t, \action_t)}{} - \frac{1}{2} \|\vu_t - \va_t  \|^2_{\vC^{-1}_{\vomega}(\id)} - \frac{1}{2} \log | \vC_{\vomega}(\id) | \notag \\
&\phantom{=} 
- \log q_{\vpsi}(\action_t|\state_t,\naction_t, \id) \Big] \Bigg] - {T\ad} \log 2\pi/2.
\end{align}
Lastly, by ignoring constants, Eq.~\eqref{eq:app_F_G_q} is equivalent to $\max_{\vphi,\vomega,\vpsi,\vtheta} \mathcal{H}(\vphi, \vomega, \vpsi, \vtheta)$, where 
\begin{align}
\mathcal{H}(\vphi, \vomega, \vpsi, \vtheta) &= \mathbb{E}_{p_{\mathrm{d}}(\state_{1:T}, \naction_{1:T}, \id)} \left[ \sum_{t=1}^T \mathbb{E}_{q_{\vpsi}(\action_t|\state_t,\naction_t, \id)} \left[ {r_{\vphi}(\state_t, \action_t)}{} - \frac{1}{2} \|\vu_t - \va_t  \|^2_{\vC^{-1}_{\vomega}(\id)} - \log q_{\vpsi}(\action_t|\state_t,\naction_t, \id) \right] \right] \notag \\
&\phantom{=} - \mathbb{E}_{\widetilde{q}_{\vtheta}(\vs_{1:T}, \va_{1:T})} \left[ \sum_{t=1}^T {r_{\vphi}(\state_t, \action_t)} - \log q_{\vtheta}(\va_t|\vs_t)  \right] + \frac{T}{2} \mathbb{E}_{p(\id)}\left[ \mathrm{Tr}(\vC_{\vomega}^{-1}(\id)\vSigma) \right]. \notag
\end{align}
This concludes the derivation of VILD. 

As mentioned, other distributions beside the Gaussian distribution can be used for $p_{\vomega}$. 
For instance, let us consider a multivariate-independent Laplace distribution: $p_{\vomega}(\naction_t|\state_t,\action_t, \id) = \Pi_{d=1}^{\ad} \frac{1}{2 c^{(d)}_k} \exp( - \| \frac{ \vu_t - \va_t}{ \vc_k } \|_1 )$, where a division of vector by vector denotes element-wise division. The Laplace distribution has heavier tails when compared to the Gaussian distribution, which makes the Laplace distribution more suitable for modeling demonstrators who tend to execute outlier actions. 
By using the Laplace distribution for $p_{\vomega}(\naction_t|\state_t,\action_t, \id)$, we obtain an objective
\begin{align}
\mathcal{H}_{\mathrm{Lap.}} &= \mathbb{E}_{p_{\mathrm{d}}(\state_{1:T}, \naction_{1:T}, \id)} \left[ \sum_{t=1}^T \mathbb{E}_{q_{\vpsi}(\action_t|\state_t,\naction_t, \id)} \left[ {r_{\vphi}(\state_t, \action_t)}{} - \Big\lVert \frac{\vu_t - \va_t}{\vc_k} \Big\rVert_1  - \log q_{\vpsi}(\action_t|\state_t,\naction_t, \id) \right] \right] \notag \\
&\phantom{=} - \mathbb{E}_{\widetilde{q}_{\vtheta}(\vs_{1:T}, \va_{1:T})} \left[ \sum_{t=1}^T {r_{\vphi}(\state_t, \action_t)} - \log q_{\vtheta}(\va_t|\vs_t)  \right] + \frac{T\sqrt{2}}{\sqrt{\pi}} \mathbb{E}_{p(\id)}\left[ \mathrm{Tr}(\vC_{\vomega}^{-1}(\id)\vSigma^{1/2}) \right]. \notag
\end{align}
We cann see that differences between $\mathcal{H}_{\mathrm{Lap}}$ and $\mathcal{H}$ are the absolute error and scaling of the trace term.

\section{Implementation details} 
\label{appendix:algo}

We implement VILD using the PyTorch deep learning framework. For all function approximators, we use neural networks with 2 hidden-layers of 100 $\mathrm{tanh}$ units, except for the Humanoid task where we use neural networks with 2 hidden-layers of 100 $\mathrm{relu}$ units.
We optimize parameters $\vphi$, $\vomega$, and $\vpsi$ by Adam with step-size $3 \times 10^{-4}$, $\beta_1 = 0.9$, $\beta_2 = 0.999$ and mini-batch size 256. To optimize the policy parameter $\vtheta$, we use trust-region policy optimization (TRPO)~\citep{SchulmanEtAl2015} with batch size 1000, except on the Humanoid task where we use soft actor-critic (SAC)~\citep{HaarnojaZAL18} with mini-batch size 256; TRPO is an on-policy RL method that uses only trajectories collected by the current policy, while SAC is an off-policy RL method that use trajectories collected by previous policies. On-policy methods are generally more stable than off-policy methods, while off-policy methods are generally more data-efficient~\citep{Gu2017}. We use SAC for Humanoid mainly due to its high data-efficiency. 
When SAC is used, we also use trajectories collected by previous policies to approximate the expectation over the trajectory density $\tilde{q}_{\vtheta}(\vs_{1:T}, \va_{1:T})$.

For the distribution $p_{\vomega}(\vu_t|\vs_t,\va_t,k) = \mathcal{N}(\vu_t, \va_t, \vC_{\vomega}(k))$, we use diagonal covariances $\vC_{\vomega}(k) = \mathrm{diag}(\vc_k)$, where $\vomega = \{ \vc_k \}_{k=1}^K$ with $\vc_k \in \mathbb{R}_{+}^{\ad}$ are parameter vectors to be learned. For the distribution $q_{\vpsi}(\va_t|\vs_t,\vu_t,\id)$, we use a Gaussian distribution with diagonal covariance, where the mean and logarithm of the standard deviation are the outputs of neural networks. Since $k$ is a discrete variable, we represent $q_{\vpsi}(\va_t|\vs_t,\vu_t,\id)$ by neural networks that have $K$ output heads and take input vectors $(\vs_t,\vu_t)$; The $k$-th output head corresponds to (the mean and log-standard-deviation of) $q_{\vpsi}(\va_t|\vs_t,\vu_t,\id)$. We also pre-train the mean function of $q_{\vpsi}(\va_t|\vs_t,\vu_t,\id)$, by performing least-squares regression for $1000$ gradient steps with target value $\vu_t$. This pre-training is done to obtain reasonable initial predictions. For the policy $q_{\vtheta}(\va_t|\vs_t)$, we use a Gaussian policy with diagonal covariance, where the mean and logarithm of the standard deviation are outputs of neural networks. We use $\vSigma = 10^{-8}\bI$ in experiments.

To control exploration-exploitation trade-off, we use an entropy coefficient $\alpha = 0.0001$ in TRPO. In SAC, we tune the value of $\alpha$ by optimization, as described in the SAC paper. Note that including $\alpha$ in VILD is equivalent to rescaling quantities in the model by $\alpha$, i.e., $\exp( {r_{\vphi}(\vs_t,\va_t)}/{\alpha})$ and $(p_{\vomega}(\vu_t|\vs_t,\va_t,\id))^{\frac{1}{\alpha}}$.
A discount factor $0 < \gamma < 1$ may be included similarly, and we use $\gamma = 0.99$ in experiments.

For all methods, we regularize the reward/discriminator function by the gradient penalty~\citep{GulrajaniAADC17} with coefficient $10$, since it was previously shown to improve performance of generative adversarial learning methods. 
For methods that learn a reward function, namely VILD, AIRL, and MaxEnt-IRL, we apply a sigmoid function to the output of reward function to control the bounds of reward function. We found that without controlling the bounds, reward values can be highly negative in the early stage of learning, which makes learning the policy by RL very challenging. 
A possible explanation is that, in MDPs with large state and action spaces, distribution of demonstrations and distribution of agent's trajectories are not overlapped in the early stage of learning. In such a scenario, it is trivial to learn a reward function which tends to positive-infinity values for demonstrations and negative-infinity values for agent's trajectories. While the gradient penalty regularizer slightly remedies this issue, we found that the regularizer alone is insufficient to prevent this scenario.


\section{Experimental Details }    \label{appendix:setting}

In this section, we describe experimental settings and data generation. We also give brief reviews of methods compared against VILD in the experiments.

\subsection{Settings and data generation}    \label{appendix:data_generation}

We evaluate VILD on four continuous control tasks from OpenAI gym platform~\citep{BrockmanEtAl2016} with the Mujoco physics simulator: HalfCheetah, Ant, Walker2d, and Humanoid. To obtain the optimal policy for generating demonstrations, we use the ground-truth reward function of each task to pre-train $\pi^\star$ with TRPO. We generate diverse-quality demonstrations by using $K=10$ demonstrators according to the graphical model in Figure~\ref{figure:pgm_cirl}. 
We consider two types of the noisy policy $p(\vu_t|\vs_t,\va_t,k)$: a Gaussian noisy policy and a time-signal-dependent (TSD) noisy policy.

\paragraph{Gaussian noisy policy.}
We use a Gaussian noisy policy $\mathcal{N}(\vu_t | \va_t, \sigma^2_{k}\bI)$ with a constant covariance. The value of $\sigma_k$ for each of the 10 demonstrators is $0.01, 0.05, 0.1, 0.25, 0.4, 0.6, 0.7, 0.8, 0.9$ and $1.0$, respectively. Note that our model assumption on $p_{\vomega}$ corresponds to this Gaussian noisy policy. Table~\ref{table:return_normal} shows the performance of demonstrators (in terms of cumulative ground-truth rewards) with this Gaussian noisy policy.

\paragraph{TSD noisy policy.}
\begin{wrapfigure}[14]{r}{0.35\textwidth}
	\vspace{-4mm}
	\centering
	\includegraphics[width=0.35\textwidth]{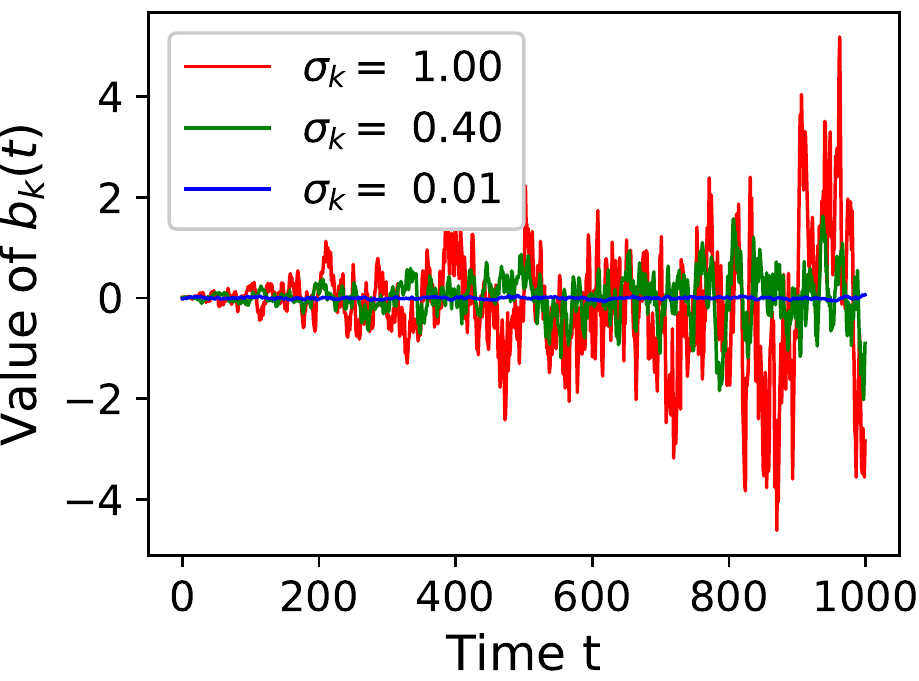}
	\caption{Samples $\vb_k(t)$ drawn from noise processes used for the TSD noisy policy.}
	\label{figure:appendix_SDNTN}
\end{wrapfigure}
To make learning more challenging, we generate demonstrations by simulating characteristics of human motor control~\citep{vanBeers2004}, where actuator noises are proportion to the magnitude of actuators, and noise's strength increases with execution time~\citep{vanBeers2004}. Specifically, we generate demonstrations using a Gaussian distribution $\mathcal{N}(\vu_t | \va_t, \diag(\vb_{k}(t) \times \| \va_t \|_1 / \ad) )$, where the covariance is proportion to the magnitude of action and depends on time step. We call this policy time-signal-dependent (TSD) noisy policy. Here, $\vb_k(t)$ is a sample of a noise process whose noise variance increases over time, as shown in Figure~\ref{figure:appendix_SDNTN}. We obtain this noise process for the $k$-th demonstrator by reversing Ornstein–Uhlenbeck (OU) processes with parameters $\theta=0.15$ and $\sigma=\sigma_k$~\citep{Uhlenbeck1930}\footnote{OU process is commonly used to generate time-correlated noises, where the noise variance decays towards zero. We reserve this process along the time axis, so that the noise variance grows over time.}. The value of $\sigma_k$ for each demonstrator is $0.01, 0.05, 0.1, 0.25, 0.4, 0.6, 0.7, 0.8, 0.9$, and $1.0$, respectively. Table~\ref{table:return_SDNTN} shows the performance of demonstrators with this TSD noisy policy. Learning from demonstrations generated by TSD is challenging; The Gaussian model of $p_{\vomega}$ cannot perfectly model the TSD noisy policy, since the ground-truth variance is a function of actions and time steps.

%
%
%

\begin{table}
	\centering
	\begin{minipage}[b]{0.47\linewidth}
		\caption{Performance of the optimal policy and demonstrators with the Gaussian noisy policy.}
		\label{table:return_normal}
		\small
		\begin{tabular}{ l || c | c | c | c}
			\hline
			$\sigma_{k}$ & Cheetah & Ant & Walker & Humanoid \\ 
			\hline \hline
			($\pi^\star$) & 4624 	& 4349 & 4963 & 5093 \\ \hline
			0.01 	& 4311  & 3985 & 4434 & 4315 \\
			0.05 	& 3978  & 3861 & 3486 & 5140 \\
			0.01 	& 4019  & 3514 & 4651 & 5189 \\
			0.25 	& 1853  & 536  & 4362 & 3628 \\
			0.40 	& 1090  & 227  & 467  & 5220 \\
			0.6  	& 567   & -73  & 523  & 2593 \\
			0.7  	& 267   & -208 & 332  & 1744 \\
			0.8  	& -45   & -979 & 283  & 735 \\
			0.9  	& -399  & -328 & 255  & 538 \\
			1.0     & -177  & -203 & 249  & 361 \\ 
		\end{tabular}
	\end{minipage}
	\hfill
	\begin{minipage}[b]{0.47\linewidth}
		\caption{Performance of the optimal policy and demonstrators with the TSD noisy policy.}
		\label{table:return_SDNTN}
		\small
		\begin{tabular}{l || c | c | c | c}
			\hline
			$\sigma_{k}$ & Cheetah & Ant & Walker & Humanoid \\ 
			\hline \hline
			($\pi^\star$) & 4624 	& 4349 & 4963 & 5093 \\ \hline
			0.01 & 4362 &	3758 & 4695 & 5130 \\
			0.05 & 4015 &	3623 & 4528 & 5099 \\
			0.01 & 3741 &	3368 & 2362 & 5195 \\
			0.25 & 1301 &	873  & 644  & 1675 \\
			0.40 & -203 &	231  & 302  & 610 \\
			0.6  & -230 &	-51  & 29   & 249 \\
			0.7  & -249 &	-37  & 24   & 221 \\
			0.8  & -416 &	-567 & 14   & 191 \\
			0.9  & -389 &	-751 & 7    & 178 \\
			1.0  & -424 &	-269 & 4    & 169 \\ 
		\end{tabular}
	\end{minipage}
	\vspace{-2mm}
\end{table}

\subsection{Comparison methods}
Here, we briefly review methods compared against VILD in our experiments. 
We firstly review online IL methods, which learn a policy by RL and require additional transition samples from MDPs. 

\vspace{-2mm}
\paragraph{MaxEnt-IRL.}
Maximum entropy IRL (MaxEnt-IRL)~\citep{ZiebartEtAl2010} is a well-known IRL method. The original derivation of the method is based on the maximum entropy principle~\citep{jaynes57} and uses a linear-in-parameter reward function: $r_{\vphi}(\vs_t,\va_t) = \vphi^\top b(\vs_t,\va_t)$ with a basis function $b$. Here, we consider an alternative derivation which is applicable to non-linear reward function~\citep{FinnLA16,FinnCAL16}. Briefly speaking, MaxEnt-IRL learns a reward parameter by minimizing a KL divergence from a data distribution $p^\star(\vs_{1:T},\va_{1:T})$ to a model $p_{\vphi}(\vs_{1:T},\va_{1:T}) = \frac{1}{Z_{\vphi}} p_1(\vs_1) \Pi_{t=1}^T p(\state_{t+1}|\vs_t,\va_t) \exp( r_{\vphi}(\vs_t,\va_t)/\alpha )$, where $Z_{\vphi}$ is the normalization term. Minimizing this KL divergence is equivalent to solving $\max_{\vphi} \mathbb{E}_{p^\star(\vs_{1:T},\va_{1:T})}\left[ \Sigma_{t=1}^T r_{\vphi}(\vs_t,\va_t) \right] - \log Z_{\vphi}$. To compute $\log Z_{\vphi}$, we can use the variational approaches as done in VILD, which leads to a max-min problem
\begin{align*}
\max_{\vphi}\min_{\vtheta} \mathbb{E}_{p^\star(\vs_{1:T},\va_{1:T})}\left[ \textstyle{\sum}_{t=1}^T r_{\vphi}(\vs_t,\va_t) \right] -  \mathbb{E}_{q_{\vtheta}(\vs_{1:T},\va_{1:T})}\left[ \textstyle{\sum}_{t=1}^T r_{\vphi}(\vs_t,\va_t) - \alpha \log q_{\vtheta}(\va_t|\vs_t) \right], \label{eq:max_ent_irl}
\end{align*}
where $q_{\vtheta}(\vs_{1:T},\va_{1:T}) = p_1(\vs_1) \Pi_{t=1}^T p(\state_{t+1}|\vs_t,\va_t) q_{\vtheta}(\va_t|\vs_t)$. The policy $q_{\vtheta}(\va_t|\vs_t)$ maximizes the  learned reward function and is the solution of IL. 

As we mentioned, the proposed model in VILD is based on the model of MaxEnt-IRL.  By comparing the max-min problem of MaxEnt-IRL and the max-min problem of VILD, we can see that the main difference are the variational distribution $q_{\vpsi}$ and the noisy policy model $p_{\vomega}$. If we assume that $q_{\vpsi}$ and $p_{\vomega}$ are Dirac delta functions: $q_{\vpsi}(\va_t|\vs_t,\vu_t,k) = \delta_{\va_t=\vu_t}$ and $p_{\vomega}(\vu_t|\va_t,\vs_t,k) = \delta_{\vu_t=\va_t}$, then the max-min problem of VILD reduces to the max-min problem of MaxEnt-IRL.  
In other words, if we assume that all demonstrators execute the same optimal policy and have an equal level of expertise, then VILD reduces to MaxEnt-IRL. 

\vspace{-2mm}
\paragraph{GAIL.} Generative adversarial IL (GAIL)~\citep{HoE16} is an IL method that perform occupancy measure matching~\citep{SyedBS08} via generative adversarial networks (GAN)~\citep{GoodfellowPMXWOCB14}. Specifically, GAIL finds a parameterized policy $\pig$ such that the occupancy measure $\rho_{\pig}(\vs,\va)$ of $\pig$ is similar to the occupancy measure $\rho_{\pi^\star}(\vs,\va)$ of $\pi^\star$. To measure the similarity, GAIL uses the Jensen-Shannon divergence, which is estimated and minimized by the following generative-adversarial training objective:
\begin{align*}
\min_{\vtheta}\max_{\vphi} \mathbb{E}_{\rho_{\pi^\star}}\left[ \log D_{\vphi}(\vs,\va) \right] + \mathbb{E}_{\rho_{\pig}}\left[ \log (1-D_{\vphi}(\vs,\va)) + \alpha \log \pi_{\vtheta}(\va_t|\vs_t) \right],
\end{align*}
where $D_{\vphi}(\vs,\va) = \frac{d_{\vphi}(\vs,\va)}{d_{\vphi}(\vs,\va) + 1}$ is called a discriminator. The minimization problem w.r.t.~$\vtheta$ is achieved using RL with a reward function $-\log(1-D_{\vphi}(\vs,\va))$.

\vspace{-2mm}
\paragraph{AIRL.}
Adversarial IRL (AIRL)~\citep{FuEtAl2018} was proposed to overcome a limitation of GAIL regarding reward function: GAIL does not learn the expert reward function, since GAIL has $D_{\vphi}(\vs,\va) = 0.5$ at the saddle point for every states and actions. To overcome this limitation while taking advantage of generative-adversarial training, AIRL learns a reward function  by solving
\begin{align*}
\max_{\vphi} \mathbb{E}_{p^\star(\vs_{1:T},\va_{1:T})}\left[ \textstyle{\sum}_{t=1}^T \log D_{\vphi}(\vs,\va) \right] + \mathbb{E}_{q_{\vtheta}(\vs_{1:T},\va_{1:T})}\left[ \textstyle{\sum}_{t=1}^T \log (1-D_{\vphi}(\vs,\va)) \right],
\end{align*}
where $D_{\vphi}(\vs,\va) = \frac{r_{\vphi}(\vs,\va)}{r_{\vphi}(\vs,\va) + q_{\vtheta}(\va|\vs)}$. 
The policy $q_{\vtheta}(\va_t|\vs_t)$ is learned by RL with a reward function $r_{\vphi}(\vs_t,\va_t)$.
\cite{FuEtAl2018} showed that the gradient of this objective w.r.t.~$\vphi$ is equivalent to the gradient of MaxEnt-IRL w.r.t.~$\vphi$. The authors also proposed an approach to disentangle reward function, which leads to a better performance in transfer learning settings. Nonetheless, this disentangle approach is general and can be applied to other IRL methods, including MaxEnt-IRL and VILD. We do not evaluate AIRL with disentangle reward function. 

We note that, based on the relation between MaxEnt-IRL and VILD, we can extend VILD to use a training procedure of AIRL. Specifically, by using the same derivation from MaxEnt-IRL to AIRL by~\cite{FuEtAl2018}, we can derive a variant of VILD which learns a reward parameter by solving $\max_{\vphi} \mathbb{E}_{p_{\mathrm{d}}(\vs_{1:T},\vu_{1:T}|k)p(k)} [ \Sigma_{t=1}^T  \mathbb{E}_{q_{\vpsi}(\va_t|\vs_t,\vu_t,k)} [ \log D_{\vphi}(\vs,\va) ] ] + \mathbb{E}_{\widetilde{q}_{\vtheta}(\vs_{1:T},\va_{1:T})} [ \Sigma_{t=1}^T \log (1-D_{\vphi}(\vs,\va)) ]$. We do not evaluate this variant of VILD in our experiment.

\vspace{-2mm}
\paragraph{VAIL.}
Variational adversarial imitation learning (VAIL)~\citep{peng2018variational} improves upon GAIL by using variational information bottleneck (VIB)~\citep{alemi2017}. VIB aims to compress information flow by minimizing a variational bound of mutual information. This compression filters irrelevant signals, which leads to less over-fitting. To achieve this in GAIL, VAIL learns the discriminator $D_{\vphi}$ by an optimization problem
\begin{align*}
\min_{\vphi, E}\max_{\beta \geq 0} &\mathbb{E}_{\rho_{\pi^\star}}\left[ \mathbb{E}_{E(\vz|\vs,\va)} \left[ -\log D_{\vphi}(\vz) \right] \right] + \mathbb{E}_{\rho_{\pig}}\left[ \mathbb{E}_{E(\vz|\vs,\va)} \left[ -\log (1-D_{\vphi}(\vz)) \right] \right] \\
&\phantom{=} + \beta \mathbb{E}_{(\rho_{\pi^\star} + \rho_{\pig})/2} \left[  \mathrm{KL}(E(\vz|\vs,\va) | p(\vz)) - I_c \right],
\end{align*}
where $\vz$ is an encode vector, $E(\vz|\vs,\va)$ is an encoder, $p(\vz)$ is a prior distribution of $\vz$, $I_c$ is the target value of mutual information, and $\beta > 0$ is a Lagrange multiplier. With this discriminator, the policy $\pig(\va_t|\vs_t)$ is learned by RL with a reward function $-\log(1-D_{\vphi}(\mathbb{E}_{E(\vz|\vs,\va)}\left[ \vz \right]))$. 

It might be expected that the compression can make VAIL robust against diverse-quality demonstrations, since irrelevant signals in low-quality demonstrations are filtered out via the encoder. However, we find that this is not the case, and VAIL does not improve much upon GAIL in our experiments. This is perhaps because VAIL compress information from both demonstrators and agent's trajectories. Meanwhile in our setting, irrelevant signals are generated only by demonstrators. Therefore, the information bottleneck may also filter out relevant signals in agent's trajectories by chance, which lead to poor performances.

\vspace{-2mm}
\paragraph{InfoGAIL.}
Information maximizing GAIL (InfoGAIL)~\citep{LiSE17} is an extension of GAIL for learning a multi-modal policy in MM-IL. The key idea of InfoGAIL is to introduce a context variable $z$ to the GAIL formulation and learn a context-dependent policy $\pig(\va|\vs,z)$, where each context represents each mode of the multi-modal policy. To ensure that the context is not ignored during learning, InfoGAIL regularizes GAIL's objective so that a mutual information between contexts and state-action variables is maximized. This mutual information is indirectly maximized via maximizing a variational lower-bound of mutual information. By doing so, InfoGAIL solves a min-max problem
\begin{align*}
\min_{\vtheta, Q}\max_{\vphi} \mathbb{E}_{\rho_{\pi^\star}}\left[ \log D_{\vphi}(\vs,\va) \right] + \mathbb{E}_{\rho_{\pig}}\left[ \log (1-D_{\vphi}(\vs,\va)) + \alpha \log \pi_{\vtheta}(\va|\vs, z) \right] + \lambda L(\pig, Q),
\end{align*}
where $L(\pig,Q) = \mathbb{E}_{p(z) \pig(\va|\vs,z)}\left[ \log Q(z|\vs,\va) - \log p(z) \right]$ is a lower-bound of mutual information, $Q(z|\vs,\va)$ is an encoder neural network, and $p(z)$ is a prior distribution of contexts. 
In our experiment, the number of context $z$ is set to be the number of demonstrators $K$. As discussed in Section~\ref{section:introduction}, if we know the level of demonstrators' expertise, then we can choose contexts that correspond to high-expertise demonstrator. In other words, we can hand-craft the prior $p(z)$ so that a probability of contexts is proportion to the level of demonstrators' expertise. For fair comparison in experiments, we do not use the oracle knowledge about the level of demonstrators' expertise, and set $p(z)$  to be a uniform distribution.

Next, we review offline IL methods. These methods learn a policy based on supervised learning and do not require additional transition samples from MDPs.

\vspace{-2mm}
\paragraph{BC.} Behavior cloning (BC)~\citep{Pomerleau88} is perhaps the simplest IL method. BC treats an IL problem as a supervised learning problem and ignores dependency between states distributions and policy. For continuous action space, BC solves a least-square regression problem to learn a parameter $\vtheta$ of a deterministic policy $\pig(\vs_t)$:
\begin{align*}
\min_{\vtheta} \mathbb{E}_{p^\star(\vs_{1:T},\va_{1:T})}\left[ \textstyle{\sum}_{t=1}^T \| \va_t -  \pig(\vs_t) \|_2^2 \right].
\end{align*} 

\vspace{-2mm}
\paragraph{BC-D.} BC with Diverse-quality demonstrations (BC-D) is a simple extension of BC for handling diverse-quality demonstrations. 
This method is based on the naive model in Section~\ref{section:model}, and we consider it mainly for evaluation purpose. BC-D uses supervised learning to learn a policy parameter $\vtheta$ and expertise parameter $\vomega$ of a model $p_{\vtheta, \vomega}( \state_{1:T}, \naction_{1:T}, k) = p(k) p(\state_1) \Sigma_{t=1}^T p(\state_{t+1}|\state_t, \naction_t) \int_{\mathcal{A}} \pi_{\vtheta}(\va_t|\vs_t) p_{\vomega}(\naction_t|\state_t,\action_t, k) \daction_t$. To learn the parameters, we minimize the KL divergence from data distribution to the model. By using the variational approach to handle integration over the action space, BC-D solves an optimization problem
\begin{align*}
\max_{\vtheta,\vomega,\vnu} \mathbb{E}_{p_\mathrm{d}(\vs_{1:T},\vu_{1:T}|k)p(k)} \left[ \textstyle{\sum}_{t=1}^T \mathbb{E}_{q_{\vnu}(\va_t|\vs_t,\vu_t,k)} 
\left[ \log \frac{ \pi_{\vtheta}(\va_t|\vs_t) p_{\vomega}(\naction_t|\state_t,\action_t, k) } { q_{\vnu}(\va_t|\vs_t,\vu_t,k)}  \right] \right],
\end{align*}
where $q_{\vnu}(\va_t|\vs_t,\vu_t,k)$ is a variational distribution with parameters $\vnu$. We note that the model $p_{\vtheta, \vomega}( \state_{1:T}, \naction_{1:T}, k)$ of BC-D can be regarded as a regression-extension of the two-coin model proposed by~\cite{RaykarYZVFBM10} for classification with noisy labels.

\vspace{-2mm}
\paragraph{Co-teaching.} Co-teaching~\citep{HanYYNXHTS18} is the state-of-the-art method to perform classification with noisy labels.
This method trains two neural networks such that mini-batch samples are exchanged under a small loss criteria. We extend this method to learn a policy by least-square regression. Specifically, let $\pi_{\vtheta_1}(\vs_t)$ and $\pi_{\vtheta_2}(\vs_t)$ be two neural networks presenting policies, and $\nabla_{\vtheta} L(\vtheta, \mathcal{B}) = \nabla_{\vtheta} \Sigma_{(\vs,\va) \in \mathcal{B}} \| \va - \pig(\vs)\|_2^2$ be gradients of a least-square loss estimated by using a mini-batch $\mathcal{B}$. The parameters $\vtheta_1$ and $\vtheta_2$ are updated by gradient iterates:
\begin{align*}
\vtheta_1 \leftarrow \vtheta_1 - \eta \nabla_{\vtheta_1} L(\vtheta_1, \mathcal{B}_{\vtheta_2} ),~\quad\quad
\vtheta_2 \leftarrow \vtheta_2 - \eta \nabla_{\vtheta_2} L(\vtheta_2, \mathcal{B}_{\vtheta_1} ).
\end{align*}
The mini-batch $\mathcal{B}_{\vtheta_2}$ for updating $\vtheta_1$ is obtained such that $\mathcal{B}_{\vtheta_2}$ incurs small loss when using prediction from $\pi_{\vtheta_2}$, i.e.,  $\mathcal{B}_{\vtheta_2} = \argmin_{\mathcal{B}'} L(\vtheta_2, \mathcal{B}')$.
Similarly, the mini-batch $\mathcal{B}_{\vtheta_1}$ for updating $\vtheta_2$ is obtained such that $\mathcal{B}_{\vtheta_1}$ incurs small loss when using prediction from $\pi_{\vtheta_1}$. For evaluating the performance, we use the first policy network: $\pi_{\vtheta_1}$.

\section{More experimental results}    \label{appendix:more_exp}

\begin{figure}[t]
	\centering
	\begin{subfigure}[b]{0.99\linewidth}
		\centering
		\includegraphics[width=0.99\linewidth]{figures/legend_iclr_crop.pdf}
	\end{subfigure}	
	
	\vspace{1mm}
	
	\centering
	\begin{subfigure}[b]{0.99\linewidth}
		\centering
		\includegraphics[width=0.24\linewidth]{figures/trpo_HalfCheetah-v2_data7_N1000_normal.pdf}
		\includegraphics[width=0.24\linewidth]{figures/trpo_Ant-v2_data7_N1000_normal.pdf}
		\includegraphics[width=0.24\linewidth]{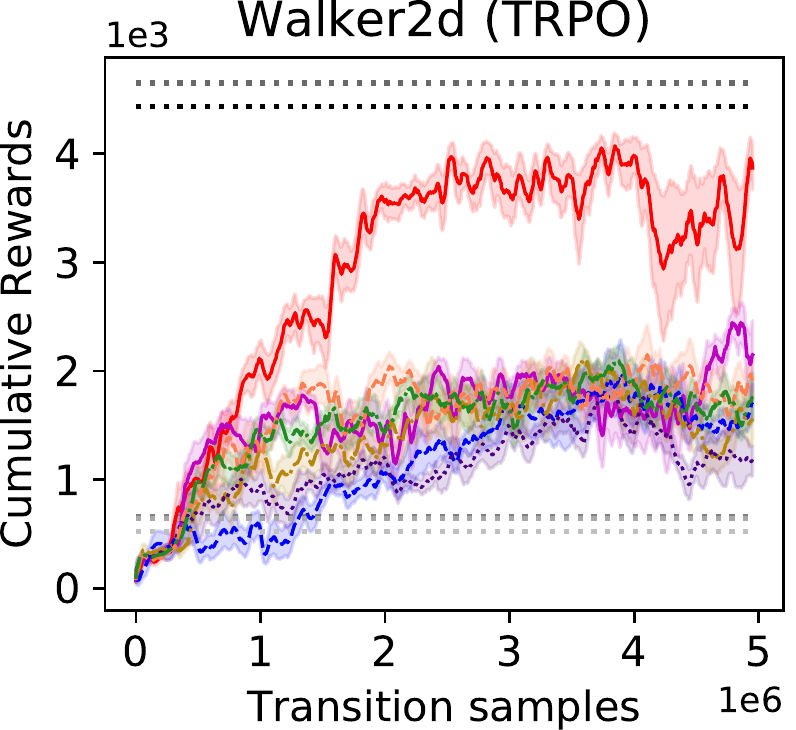}
		\includegraphics[width=0.24\linewidth]{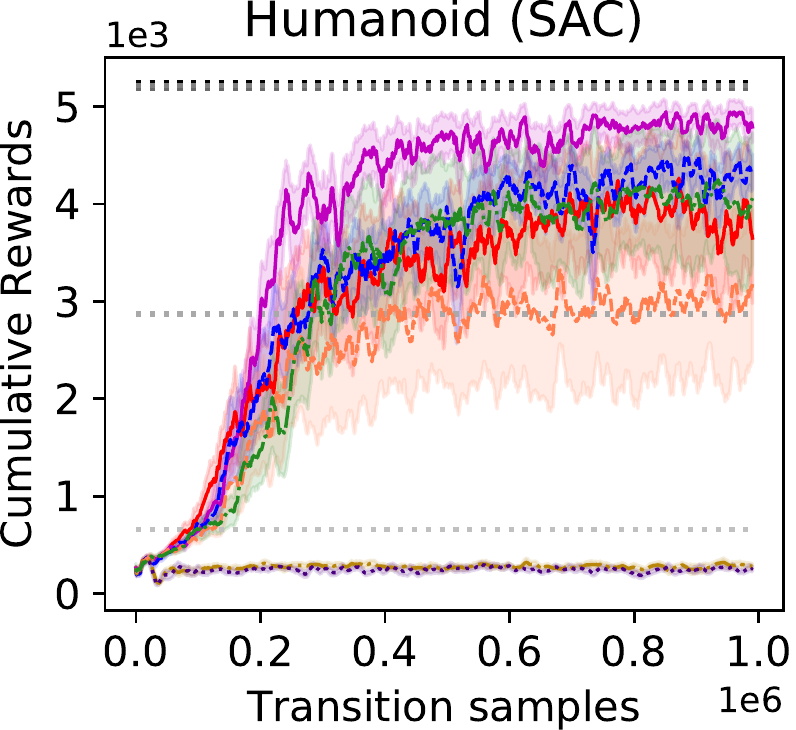}
		\hfill
		\subcaption{Performan of online IL methods when demonstrations are generated using Gaussian noisy policy.}
		\label{figure:app_exp_trpo_normal}
	\end{subfigure}	
	\hfill
	
	\vspace{1mm}
	
	\begin{subfigure}[b]{0.99\linewidth}
		\centering
		\includegraphics[width=0.24\linewidth]{figures/trpo_HalfCheetah-v2_data7_N1000_SDNTN.pdf}
		\includegraphics[width=0.24\linewidth]{figures/trpo_Ant-v2_data7_N1000_SDNTN.pdf}
		\includegraphics[width=0.24\linewidth]{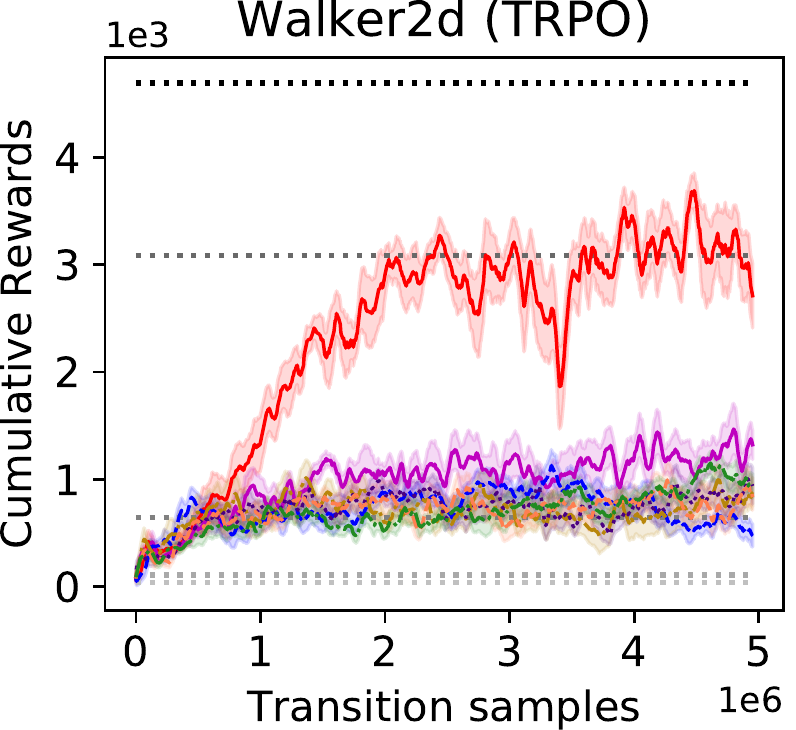}
		\includegraphics[width=0.24\linewidth]{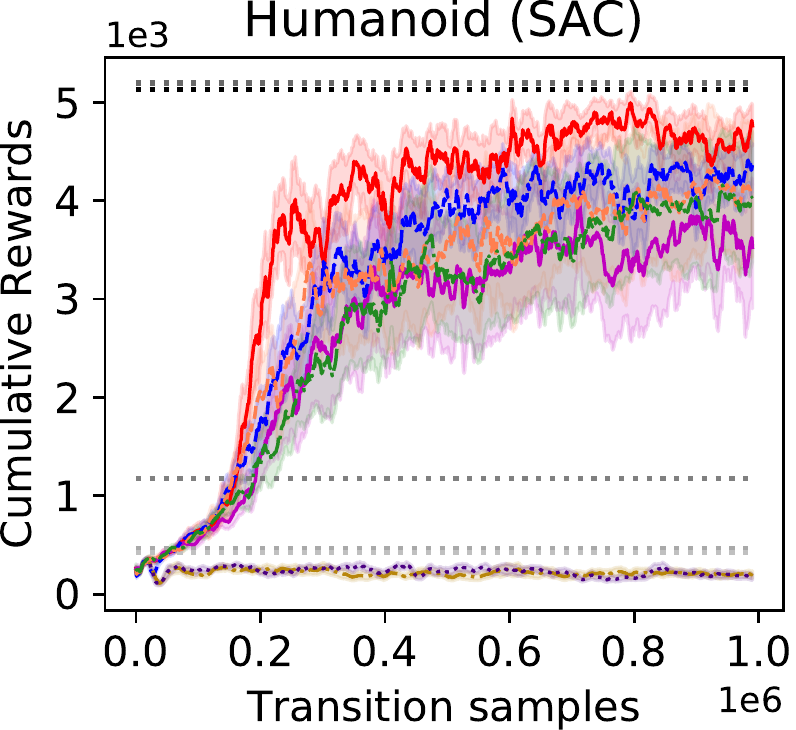}
		\hfill
		\subcaption{Performan of online IL methods when demonstrations are generated using TSD noisy policy.}
		\label{figure:app_exp_trpo_SDNTN}
	\end{subfigure}	
	\caption{Performance averaged over 5 trials of online IL methods against the number of transition samples. Horizontal dotted lines indicate performance of $k=1,3,5,7,10$ demonstrators.}	
	\label{figure:exp_app_online}
	
	\vspace{2mm}
	
	\centering
	\begin{subfigure}[b]{0.90\linewidth}
		\centering
		\includegraphics[width=0.85\linewidth]{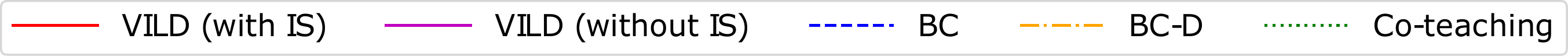}
	\end{subfigure}	
	
	\vspace{1mm}
	
	\centering
	\begin{subfigure}[b]{0.99\linewidth}
		\centering
		\includegraphics[width=0.24\linewidth]{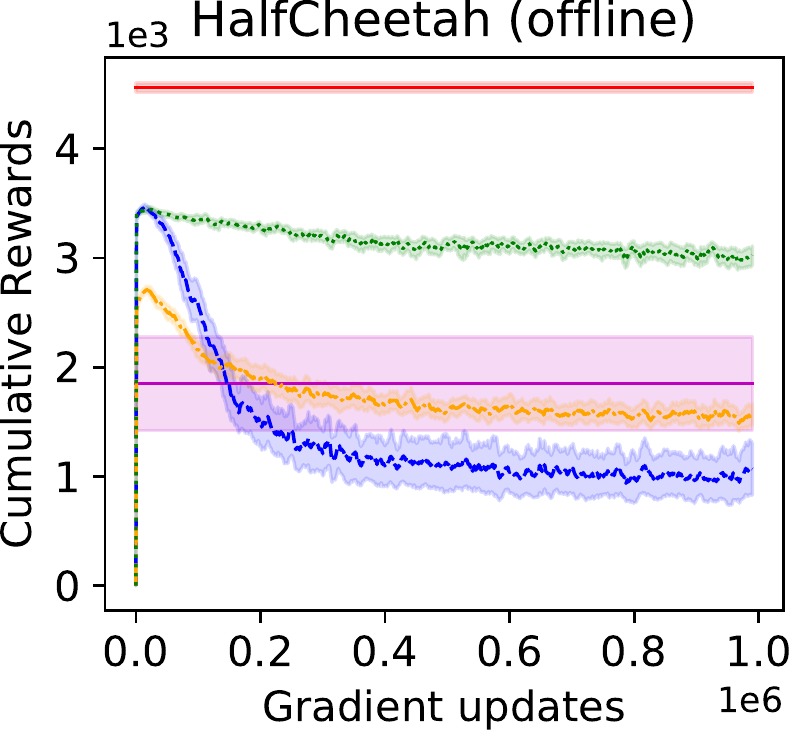}~
		\includegraphics[width=0.24\linewidth]{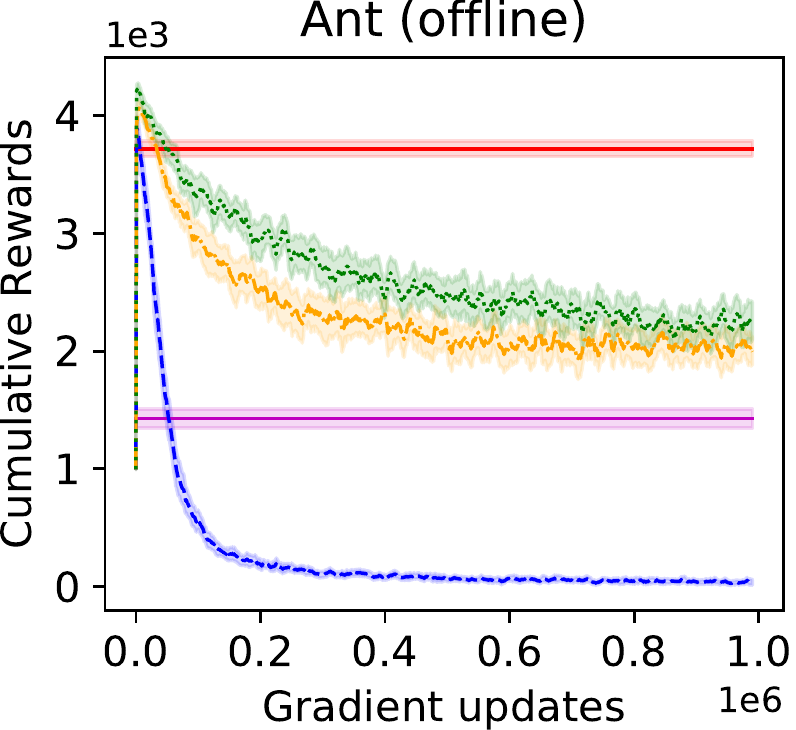}~
		\includegraphics[width=0.24\linewidth]{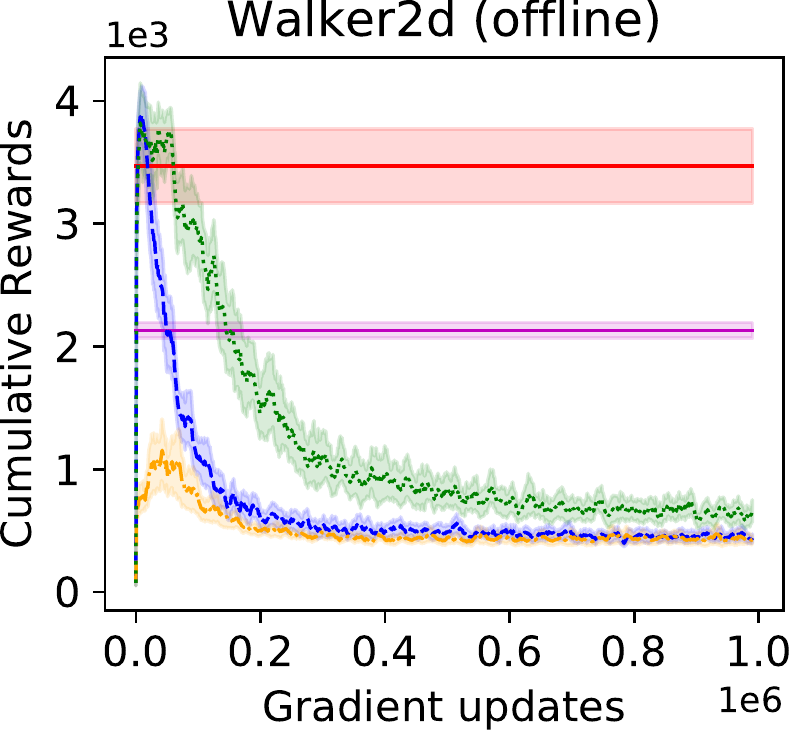}~
		\includegraphics[width=0.24\linewidth]{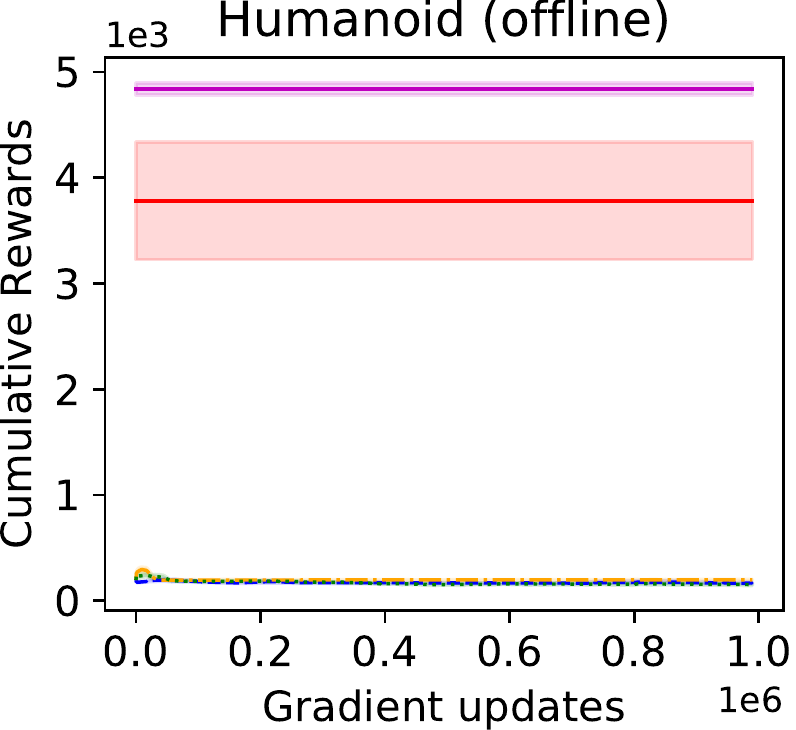}~
		\hfill
		\subcaption{Performan of offline IL methods  when demonstrations are generated using Gaussian noisy policy.}
		\label{figure:app_exp_trpo_normal_offline}
	\end{subfigure}	
	\hfill
	
	\vspace{1mm}
	
	\begin{subfigure}[b]{0.99\linewidth}
		\centering
		\includegraphics[width=0.24\linewidth]{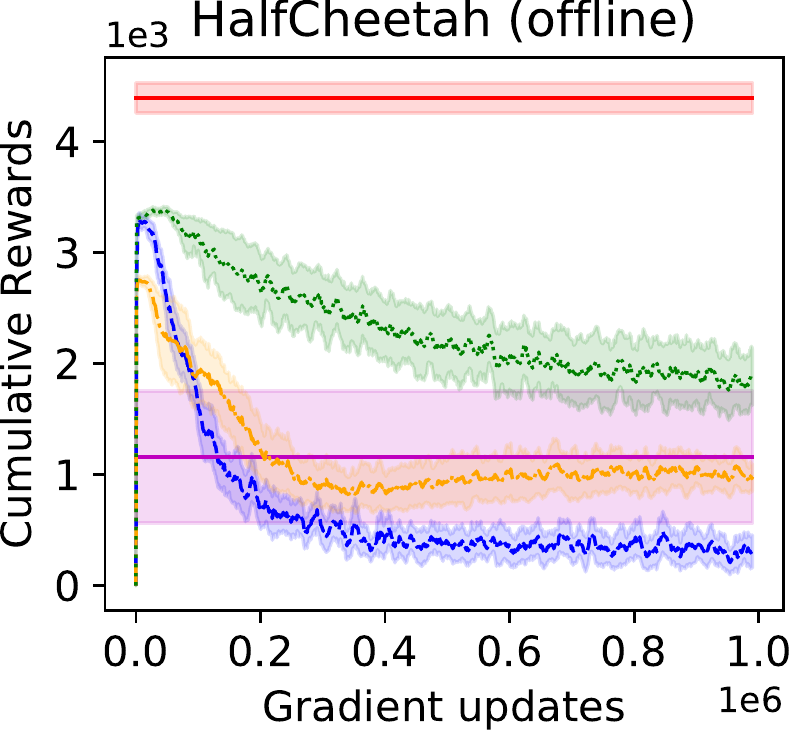}~
		\includegraphics[width=0.24\linewidth]{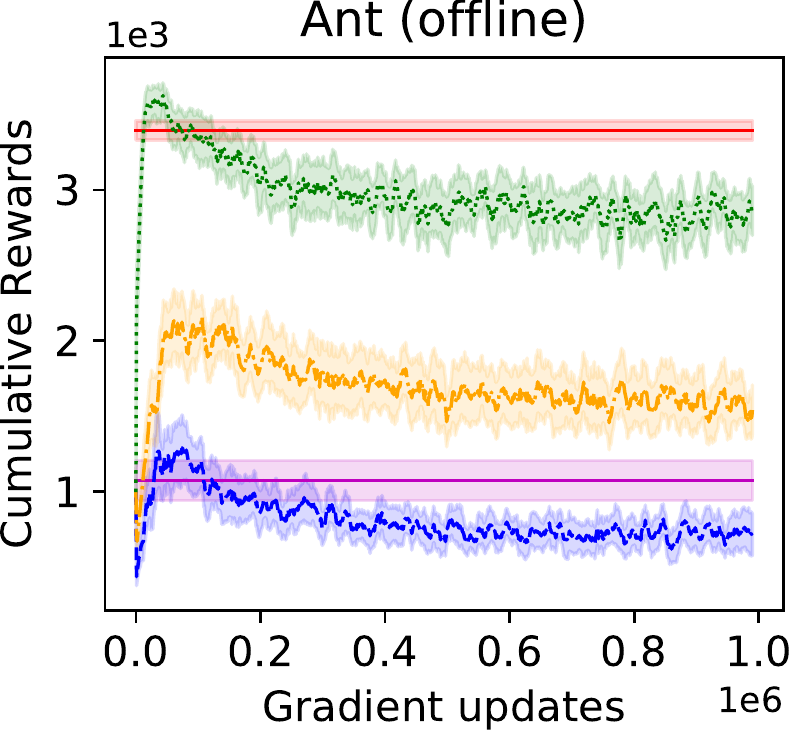}~
		\includegraphics[width=0.24\linewidth]{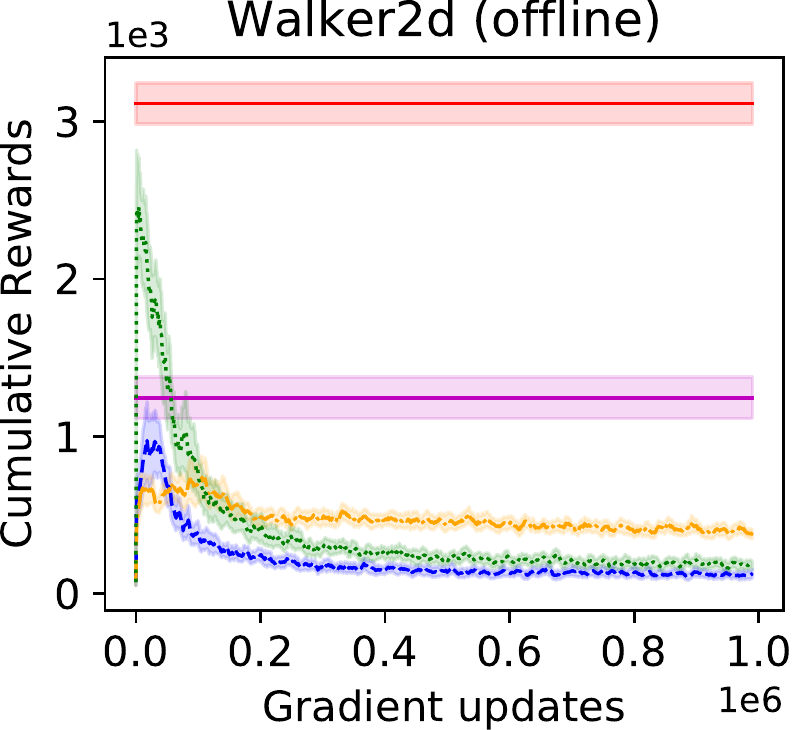}~
		\includegraphics[width=0.24\linewidth]{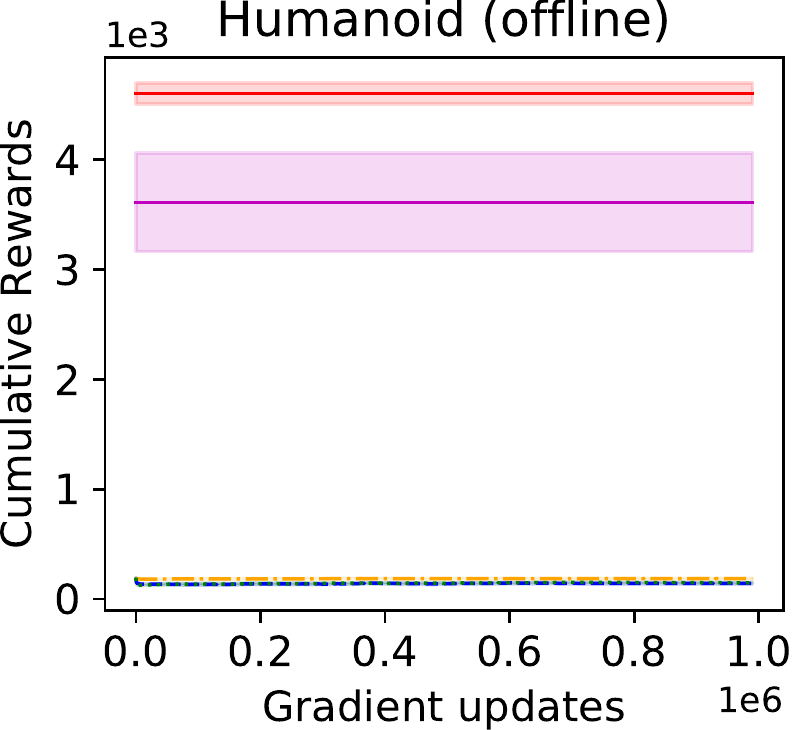}~
		\hfill
		\subcaption{Performan of offline IL methods  when demonstrations are generated using TSD noisy policy.}
		\label{figure:app_exp_trpo_SDNTN_offline}
	\end{subfigure}	
	
	\caption{Performance averaged over 5 trials of offline IL methods against the number of gradient update steps. For VILD with and without IS, we report the final performance in Table~\ref{table:performance}. }	
	\label{figure:exp_app_offline}
	
	\vspace{2mm}
	
	\centering
	\includegraphics[width=0.24\linewidth]{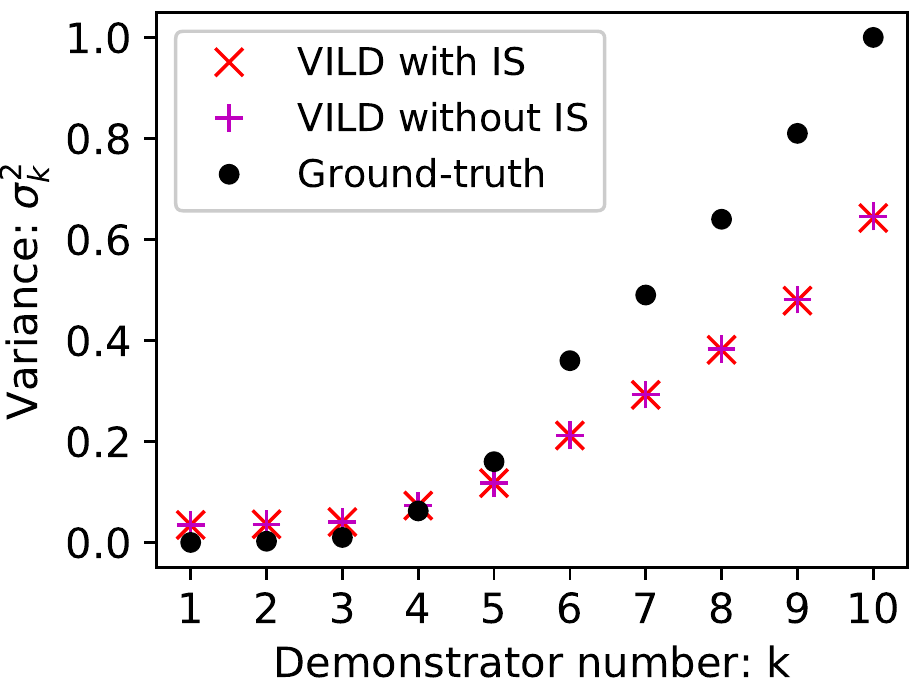}~~
	\includegraphics[width=0.24\linewidth]{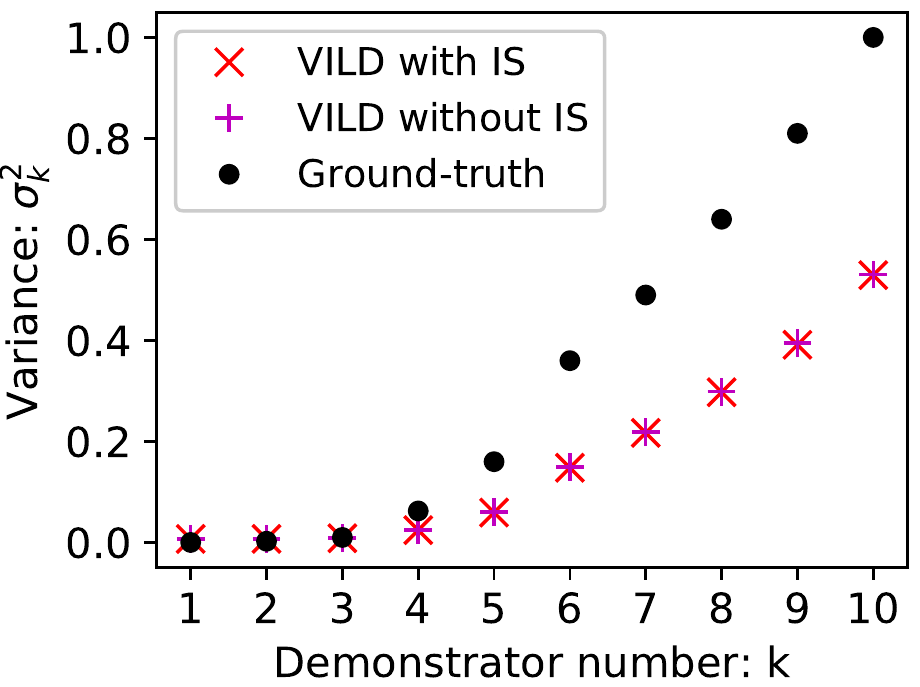}~~
	\includegraphics[width=0.24\linewidth]{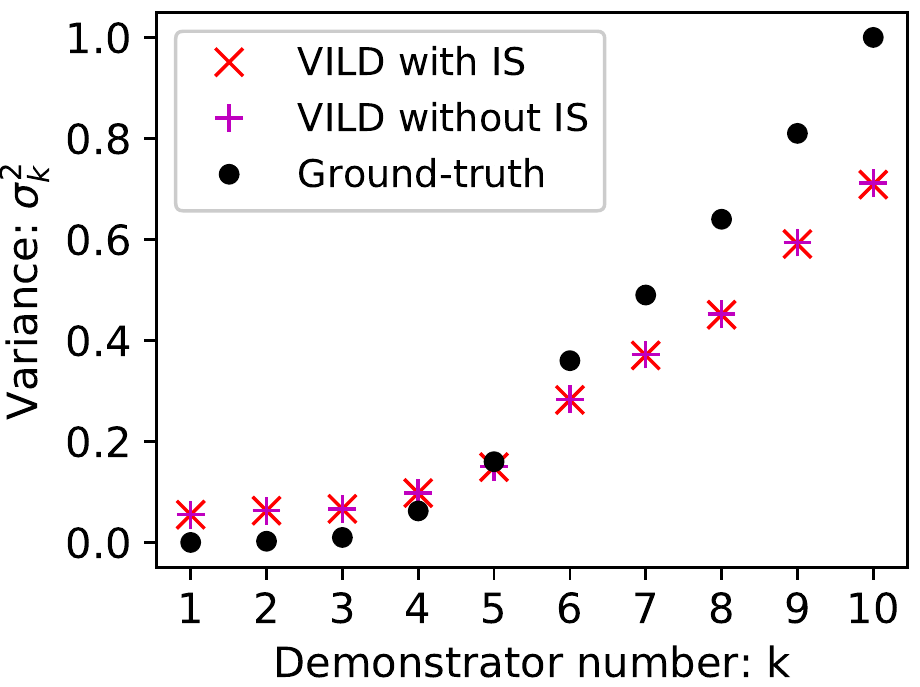}~~
	\includegraphics[width=0.24\linewidth]{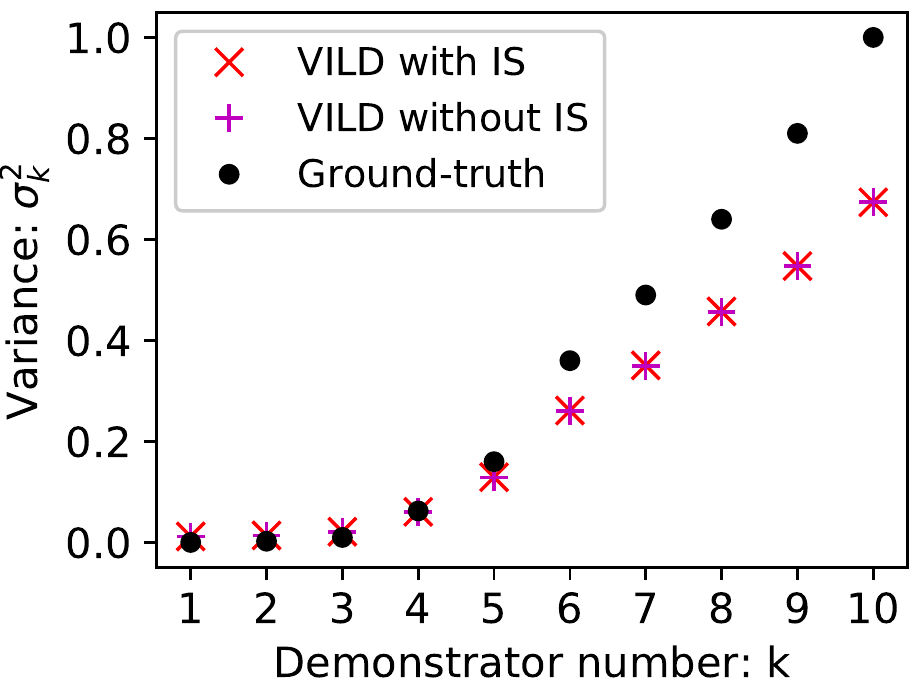}~~
	\hfill		
	\caption{Expertise parameters $\vomega = \{ \vc_k \}_{k=1}^K$ learned by VILD and the ground-truth $\{ \sigma_k^2 \}_{k=1}^K$ for the Gaussian noisy policy. For VILD, we report the value of $\| \vc_k \|_1 / \ad$.}	
	\label{figure:exp_app_online_expertise}
\end{figure}

\vspace{-2mm}
\paragraph{Results against online IL methods.}
Figure~\ref{figure:exp_app_online} shows the learning curves of VILD and existing online IL methods against the number of transition samples.
It can be seen that for both types of noisy policy, VILD with and without IS outperform existing methods overall, in terms of both final performance and data-efficiency. 

\vspace{-2mm}
\paragraph{Results against offline IL methods.}
Figure~\ref{figure:exp_app_offline} shows learning curves of offline IL methods, namely BC, BC-D, and Co-teaching. 
For comparison, the figure also shows the final performance of VILD with and without IS, according to Table~\ref{table:performance}.
We can see that these offline methods do not perform well, especially on the high-dimensional Humanoid task. The poor performance of these methods is due to the issues of compounding error and low-quality demonstrations. Specifically, BC performs the worst, since it suffers from both issues. Still, BC may learn well in the early stage of learning, but its performance sharply degrades, as seen in Ant and Walker2d. This phenomena can be explained as an empirical effect of {memorization} in deep neural networks~\citep{ArpitJBKBKMFCBL17}. Namely, deep neural networks learn to remember samples with simple patterns first (i.e., high-quality demonstrations from experts), but as learning progresses the networks overfit to samples with difficult patterns (i.e., low-quality demonstrations from amateurs). Co-teaching is the-state-of-the-art method to avoid this effect, and we can see that it performs overall better than BC. Meanwhile, BC-D, which learns the policy and level of demonstrators' expertise, also performs better than BC and is comparable to Co-teaching. However, due to the presence of compounding error, the performance of Co-teaching and BC-D is still worse than VILD with IS.

%
%
%

\vspace{-2mm}
\paragraph{Accuracy of estimated expertise parameter.}
Figure~\ref{figure:exp_app_online_expertise} shows the estimated parameters $\vomega = \{ \vc_k \}_{k=1}^K$ of $\mathcal{N}(\vu_t| \va_t, \mathrm{diag}(\vc_k))$ and the ground-truth variance  $\{ \sigma_k^2 \}_{k=1}^K$ of the Gaussian noisy policy $\mathcal{N}(\vu_t| \va_t, \sigma_k^2\bI)$. The results show that VILD learns an accurate ranking of the variance compared to the ground-truth. The values of these parameters are also quite accurate compared to the ground truth, except for demonstrators with low-levels of expertise. A possible reason for this phenomena is that low-quality demonstrations are highly dissimilar, which makes learning the expertise more challenging. We can also see that the difference between the parameters of VILD with IS and VILD without IS is small and negligible.

\end{document}